# SHAMS
# An Audio-Grounded Pronunciation Benchmark for Levantine Arabic

**Ben Sapirstein**
Reichman University

**Roy Mattar**
Independent Researcher

**Guy Mor-Lan**
The Hebrew University of Jerusalem

**Ahlam Mohamed**
Tel Aviv University

**Letizia Cerqueglini**
Tel Aviv University

**Morris Alper**
University of Miami

## Abstract

Levantine Arabic (LA) is spoken by tens of millions of people, creating a pressing need for shared benchmarks to evaluate LA speech-language technologies. Evaluating such technology is particularly challenging given LA's internal diversity and its opaque and non-standardized orthography. We present **SHAMS** (*SHami Annotated Multi-dialect Speech*), a benchmark comprising 1,300 utterances drawn from open audio corpora, balanced across five LA varieties (Urban and Rural Palestinian, and Urban Jordanian, Lebanese, and Syrian). Each utterance is represented across four aligned tiers: audio, unvocalized orthography, diacritized text, and phonetic transcription. This structure supports evaluation of various downstream tasks such as diacritization, grapheme-to-phoneme conversion, automatic speech recognition, and audio-to-phoneme, grounded in audio and stratified by variety. We benchmark open and proprietary models across these tasks to demonstrate the utility of this benchmark for measuring progress across LA. We release SHAMS at `https://shams-nlp.github.io`.

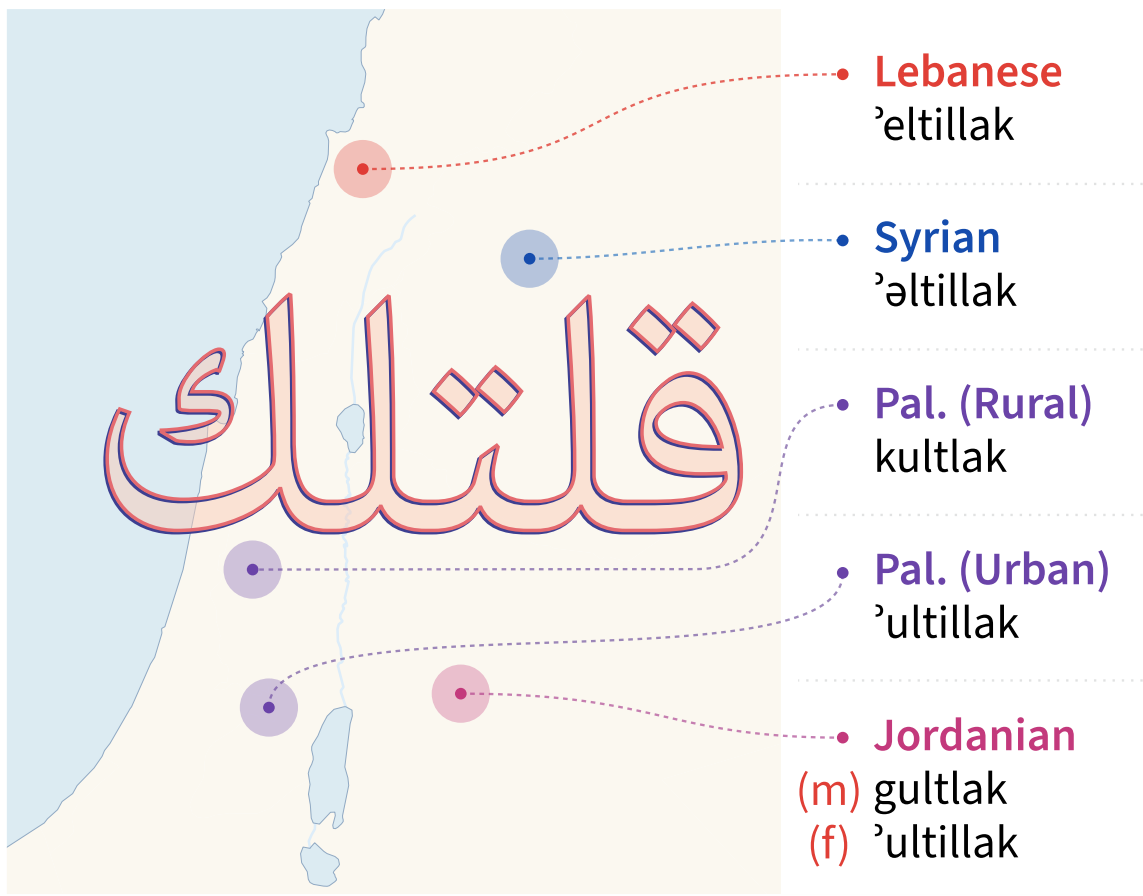

Figure 1: **SHAMS** includes audio and text for the Levantine Arabic varieties shown in the map above. Above, we illustrate different pronunciations of the word قلتلك (colloquial spelling, meaning *I told you*) to show the importance of our phonetic annotations; these pronunciation differences are not indicated in Arabic orthography, even with conventional diacritics.

## 1 Introduction

As speech technology improves for applications such as assistive technology and conversational AI, progress depends on shared evaluation benchmarks. However, these remain unevenly developed across the world's languages. Levantine Arabic (LA), spoken by tens of millions of people across Palestine, Israel, Jordan, Lebanon, and Syria (Brustad and Zuniga, 2019), poses particular challenges: it shows significant internal diversity across regions and speakers, and its written form is non-standardized and leaves key phonetic features unspecified. As a result, there is a pressing need for shared evaluation benchmarks for LA speech-language tasks.

Most Arabic-speaking communities exist in a state of *diglossia*, where regional varieties such as LA coexist with Modern Standard Arabic (MSA), a codified register learned through formal schooling and used in writing, news, and oratory, but not natively spoken in everyday conversation (Ferguson, 1959). Existing work largely focuses on MSA, with dialectal varieties significantly underrepresented in both corpora and models (Djanibekov et al., 2025; Lau et al., 2025). Recent work finds that even LLMs that can understand dialectal Arabic tend to default to MSA in generation (Robinson et al., 2025).

This is consistent with the well-known Arabic sociolinguistic trend of dialect leveling—the reduction of linguistic differences between varieties through contact and standardization (Al-Wer, 2007; Al-Wer et al., 2022).

Arabic is written in an *abjad*, a script type that represents consonants but omits short vowels and other phonetic features such as gemination (Daniels, 1990). Consider the word بقدر (*bqdr*). Without diacritics, it is ambiguous between three readings: بِقَدْر / بِقَدَر (*bi-qadr / bi-qadar*; ‘according to’); بَقْدِر (*baqdir*; ‘I can’); and بِقَدِّر (*bi-qaddir*; ‘I estimate’). Even with diacritics added, بَقْدِر alone still underspecifies the regional pronunciation—across Palestinian varieties, ق (/q/) surfaces as a glottal stop [ʔ] in urban speech (*ba’dir*), as a voiceless velar [k] in rural speech (*bakdir*), and as a voiced velar [g] in Bedouin speech (*bagdir*). Figure 1 illustrates this diversity in pronunciation.

Arabic dialectology directly studies such regional pronunciations, using phonetic transcriptions of recorded speech (Henkin, 2010; Cerqueglini, 2022, 2024). When transcripts are only available as unvocalized Arabic text, these pronunciation differences are obscured and cannot be studied. However, producing phonetic transcriptions by hand is slow, and has been described as a major bottleneck for data collection in dialectology (Kuparinen, 2025). Automating this transcription could relieve the bottleneck, but for LA there has been no way to measure how well current systems perform the task.

Bridging the gap between written form and pronunciation involves several key tasks, including diacritization (restoring vowel and gemination marks), grapheme-to-phoneme conversion (G2P; mapping orthography to phonetic transcription), automatic speech recognition (ASR; transcribing audio to text), and audio-to-phoneme conversion (A2P; transcribing audio to phonetic transcription). Diacritization has been studied extensively for MSA (Mohamed and Mubarak, 2025), with more limited work on G2P and Arabic dialects (Masmoudi et al., 2019; Lodagala et al., 2025). Dialectal Arabic ASR has been advanced by recent audio corpora (Talafha et al., 2024; Sullivan et al., 2026). These efforts address individual tasks in isolation, and neither source corpus provides diacritics, a phonetic tier, or variety-level stratification; our novel contribution is an evaluation that ties diacritization, G2P, ASR, and A2P together over the same utterances, grounded in audio and stratified by LA variety.

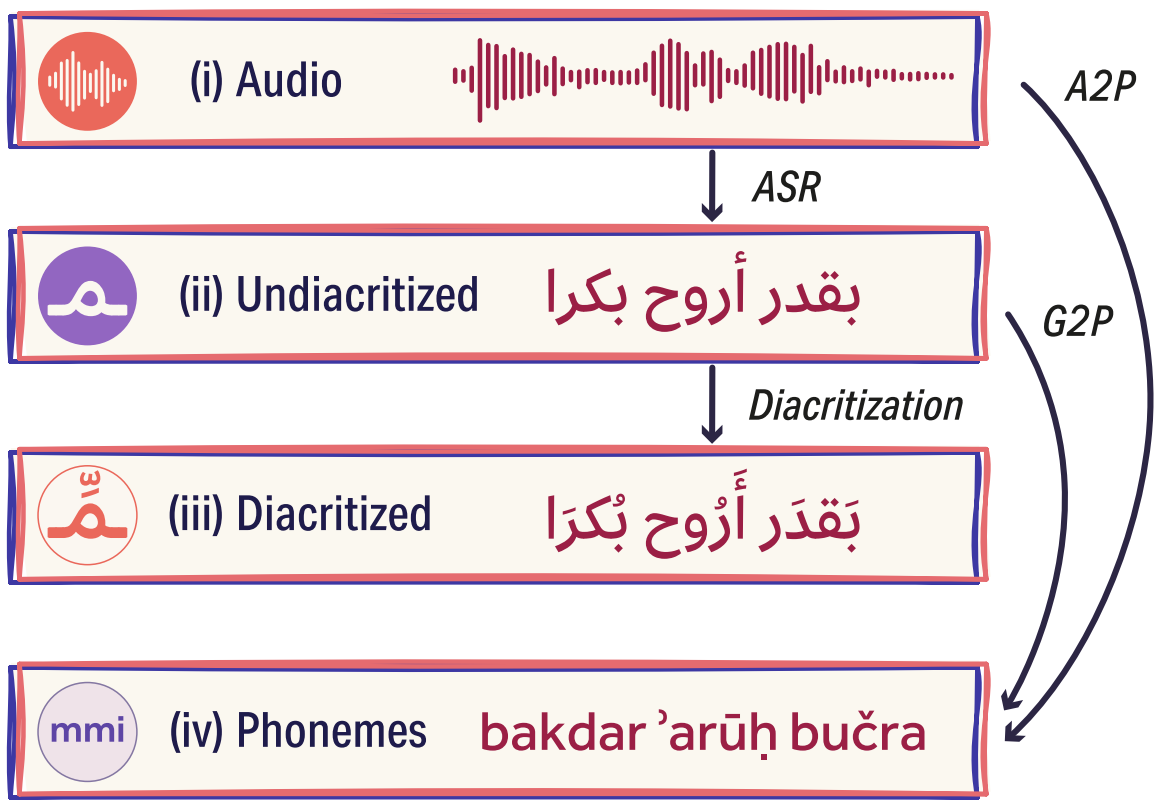

Figure 2: **SHAMS contents and benchmark tasks.** Each utterance in SHAMS comprises four aligned tiers (§2.2): (i) audio, (ii) undiacritized text, (iii) diacritized text and (iv) phonetic transcription. Arrows denote evaluation tasks; e.g., G2P predicts the phonetic transcription from unvocalized text.

To address this, we present **SHAMS** (*SHami Annotated Multi-dialect Speech*), an open benchmark for spoken LA. SHAMS comprises 1,300 utterances drawn from open audio corpora, balanced across varieties by word count, each represented across four aligned tiers (Figure 2). SHAMS is designed as an evaluation benchmark to test the robustness of LA systems beyond their training distribution. Recent work on other Semitic languages has shown that small expert-annotated evaluation benchmarks for tasks like G2P and ASR can drive substantial progress, including enabling training on large unlabeled corpora (Kolani et al., 2025; Melichov et al., 2026). We release SHAMS at the URLs below.[1]

To summarize, our key contributions are:

- **A balanced spoken corpus** of 1,300 utterances drawn from open audio sources, covering five Levantine varieties
- **Multi-tier annotations** aligned to each utterance consisting of diacritized Arabic and phonetic transcription, thus enabling joint evaluation of diacritization, G2P, ASR, and A2P.
- **Baseline results** across open and proprietary models, stratified by variety, demonstrating the benchmark's utility for distinguishing model performance across tasks and varieties.

[1]Data: https://huggingface.co/datasets/shams-nlp/shams. Evaluation code and model outputs: https://github.com/bensapirstein/shams.

## 2 SHAMS

### 2.1 Sources and Contents

SHAMS comprises 1,300 utterances totaling roughly 15,500 words and just over two hours of speech, spread across five Levantine varieties: Palestinian Urban, Palestinian Rural, Jordanian, Lebanese, and Syrian. Besides Palestinian Rural, all varieties in the benchmark are urban. The benchmark is roughly balanced by word counts, with about 3,100 words per variety, and contains a mix of speaker genders. Detailed statistics are provided in Table 1.

The benchmark is built on the Levantine portions of two open multi-dialectal Arabic speech corpora. From *Casablanca* (Talafha et al., 2024) we take scripted dialectal television dialogue; from the *LangMedia* archive (Five College Center for World Languages, n.d.), documentary interviews and spontaneous conversation on topics such as food, family, and daily life. The data features speakers of different generations and expressive styles who routinely mix registers and incorporate loanwords from contact languages.

We retain *Casablanca*'s existing utterance segmentation and identifiers, and segment the *LangMedia* recordings into short utterances following Talafha et al. (2024), excluding segments with fewer than three words or heavy background noise. We exclude designated training-split data, as SHAMS is designed solely for evaluation. Where original transcripts do not match the accompanying audio, we correct them manually.

As part of preprocessing we normalize Arabic Unicode across all text tiers, folding presentation forms, removing invisible controls and *tatweel*, and applying canonical composition so that diacritic sequences compare consistently. We apply no orthographic normalization at this stage: for example, *hamza* and *alif* variants, *tāʾ marbūṭa* and *alif maqṣūra* are left as originally transcribed.

### 2.2 Benchmark Structure

As shown in Figure 2, each utterance in SHAMS includes four aligned tiers: (i) the **audio** segment; (ii) **unvocalized Arabic orthography**, the consonantal text as conventionally written, corrected when needed to match the recording; (iii) **fully diacritized Arabic**, the same text with conventional diacritics matching the pronunciation in the audio; and (iv) a **phonetic transcription** in a Semitic transcription system.

| Variety | Source | Utt. | Words | M/F % | Min. |
|---|---|---|---|---|---|
| Pal./Urban | Casablanca | 22 | 267 | 49.8/50.2 | 1.9 |
| | LangMedia | 208 | 2,908 | 48.5/51.5 | 23.1 |
| | both | 230 | 3,175 | 48.6/51.4 | 25.0 |
| Pal./Rural | Casablanca | 234 | 3,153 | 69.0/31.0 | 21.8 |
| Jordanian | Casablanca | 137 | 1,451 | 60.0/40.0 | 10.5 |
| | LangMedia | 168 | 1,722 | 35.0/65.0 | 15.0 |
| | both | 305 | 3,173 | 46.5/53.5 | 25.5 |
| Lebanese | LangMedia | 316 | 3,008 | 49.9/50.1 | 29.1 |
| Syrian | LangMedia | 215 | 2,997 | 50.0/50.0 | 26.7 |
| **Total** | | **1,300** | **15,506** | **52.8/47.2** | **128.2** |

Table 1: SHAMS composition by variety and source: utterances, words, the male/female share of words in per cent, and duration in minutes. Palestinian Urban and Jordanian each draw on both source corpora; Palestinian Rural comes from Casablanca only, and Lebanese and Syrian from LangMedia only.

### 2.3 Annotation

**Diacritics.** Diacritization, also called vocalization, is the process of adding conventional pronunciation markings to consonantal Arabic text. We annotate five categories. *Fatḥa*, *kasra*, and *damma* mark the three short vowels /a/, /i/, and /u/; *shadda* marks gemination of the consonant it sits on; and *tanwīn* marks nunation, a final /n/ added to a short vowel. We do not annotate *sukūn*, the explicit mark of a vowelless consonant.

**Phonetic transcription.** Even fully diacritized orthography leaves dialectal pronunciation underspecified (§1): for example, the realization of ق is not recoverable from the script. We therefore add a phonetic transcription tier, following the Semitic transcription system adopted by the *Encyclopedia of Arabic Language and Linguistics* (Versteegh, 2006–2009) and used in Arabic dialectological scholarship (Cerqueglini, 2022, 2024; Henkin, 2010; Shahin, 2011). While this notation follows Arabic dialectological conventions, in principle it maps to the International Phonetic Alphabet (IPA) by a deterministic character correspondence. Table 2 illustrates phenomena that surface in the phonetic transcription but not in the orthographic tier.

**Annotation principles and procedure.** The data exhibits substantial phonetic variability across and within dialects (Appendix A). Our guiding principle is to document this diversity as-is, rather than normalizing to canonical forms. The diacritized and phonetic tiers record the realization heard in each recording, so the same orthographic word may carry different diacritics and phonetic transcriptions across utterances, even within the same dialect.

| Example 1: Code-switching and *imāla* |
|---|
| selle d’agneau كمان عندك مثل |
| ***kamēn*** *ʿandek* ***mitl sel danyō*** |
| **Example 2: Register mixing** |
| من خبرتك الطويلة بالسوق، قديش تأثر الاقتصاد السوري بالأزمة هي؟ |
| *min ḫibritak iṭ-ṭawīle* ***bis-sūʾ****,* ***ʾaddēš taʾassor il-iqtiṣād*** *is-sūri bi-l-ʾazme hiyye?* |

Table 2: **Necessity of phonetic transcription.** Two *LangMedia* utterances (Lebanese, top; Syrian, bottom) show phenomena in SHAMS that are visible only in the phonetic tier. In Example 1, the French loan is written as pronounced (*sel danyō*), مثل shifts the fricative ث to plosive [t] and كمان carries *imāla* (*kamēn*). In Example 2, the same speaker realizes ق as [q] in the MSA-register noun *il-iqtiṣād* and as [ʔ] in the everyday *bis-sūʾ* and *ʾaddēš*, and shifts the interdental ث to a sibilant in *taʾassor*.

Both pronunciation tiers were produced semi-automatically. For 520 initial utterances, phonetic transcriptions were produced by hand and aligned to the Arabic orthography to place diacritics. For the remaining 780 utterances, Levanti (Mor-Lan, 2024) produced an automatic first pass over both tiers, which annotators then corrected. Consonantal spelling was also corrected against the audio during annotation. Annotators worked against the audio throughout, and every utterance was reviewed by an author. Because Levanti is also evaluated as a baseline (§3.1), we mark it as potentially in-distribution in result tables and we note that 161 utterances (12.4%) carry diacritics identical to its output after human review. Full annotation conventions are provided in Appendix A.

**Annotators.** Diacritics were added by three native speakers of Palestinian Arabic, led by a PhD student specializing in Arabic dialectology. Phonetic transcriptions were produced by a co-author with a PhD in Semitic linguistics and publications in Arabic dialectology.

### 2.4 Annotation Reliability

A second native speaker of Palestinian Arabic independently re-annotated a stratified subset of 150 utterances, 30 per variety, under the same audio-guided protocol. We measure agreement over the diacritic slot, one per Arabic letter, giving 7,290 paired observations. Agreement is 93.2%, with Cohen’s $\kappa = 0.898$ (Table 3). Intervals are 95% confidence intervals from a cluster bootstrap resampling full utterances, in order to account for the non-independence of slots within an utterance.

| | Utt. | Slots | Agree | $\kappa$ [95% CI] |
|---|---|---|---|---|
| **Overall** | 150 | 7,290 | 93.2% | 0.898 [0.886, 0.909] |
| Pal./Urban | 30 | 2,237 | 93.1% | 0.898 [0.881, 0.917] |
| Pal./Rural | 30 | 1,484 | 95.4% | 0.930 [0.913, 0.944] |
| Jordanian | 30 | 1,008 | 91.3% | 0.867 [0.823, 0.901] |
| Lebanese | 30 | 1,078 | 93.2% | 0.896 [0.863, 0.927] |
| Syrian | 30 | 1,483 | 92.4% | 0.886 [0.861, 0.912] |
| Palestinian | 60 | 3,721 | 94.0% | 0.911 [0.898, 0.924] |
| Non-Palestinian | 90 | 3,569 | 92.4% | 0.883 [0.865, 0.901] |

Table 3: Inter-annotator agreement on the diacritization subset. The unit is the diacritic slot: one per Arabic letter, labelled with the marks it carries. Words whose consonantal skeletons differ between the two annotators are an orthographic divergence rather than a diacritic disagreement and are excluded (142 of 1,881 words). Intervals are cluster bootstraps resampling utterances.

**Protocol.** The second annotator was shown the first annotator’s consonantal skeleton, so the comparison measures diacritic reliability only. The subset and both annotators’ labels are released with the benchmark.

**Cross-variety agreement.** Agreement is high ($\kappa \geq 0.86$) across all varieties. All diacritization annotators are native Palestinian Arabic speakers; agreement is slightly higher on Palestinian material ($\kappa = 0.911$) than elsewhere ($0.883$), a difference of $0.027$ (95% CI $[0.006, 0.050]$), consistent with the audio-grounded protocol reducing reliance on annotators’ native-dialect intuitions.

**Disagreements.** Most disagreement is about whether a vowel is marked at all (عَلَى / عَلى), then whether to geminate (بَسّ / بَس), and then which vowel to indicate (جِدًّا / جدًّا). A separate set of words differs in spelling rather than diacritics, mostly *hamza* seating (إنو / انو), and sometimes a dialectal form (متل / مثل).

### 2.5 Release and Licensing

SHAMS annotations are released under terms consistent with each source corpus’s license. Utterances sourced from LangMedia and accompanying annotations are released in compliance with LangMedia’s CC BY-SA license. Annotations linked to Casablanca utterances are released with identifiers without redistributing audio; written correspondence with its authors confirmed this format is consistent with their CC BY-NC-ND license. Full license terms are provided with the released dataset.

## 3 Experiments and Results

We evaluate models on four tasks enabled by the SHAMS tier structure: diacritization, grapheme-to-phoneme (G2P) conversion, automatic speech recognition (ASR), and audio-to-phoneme (A2P) conversion. We test both dedicated systems and general-purpose LLMs. We report Diacritic Error Rate (DER) for diacritization, computed as in Catt (Alasmary et al., 2024), which aligns the diacritic sequence of each letter and word boundary, Character Error Rate (CER) for remaining tasks, as well as Word Error Rate (WER) for ASR. Results are reported in aggregate and stratified by variety; per-source results for the two varieties drawn from both corpora are in §B.2.

**Scoring.** Every score is computed after an evaluation-time normalization, applied symmetrically to references and hypotheses, at two levels. The *strict* tier scores the output as written; the *neutral* tier additionally folds spelling and notation that carry no phonemic contrast in Levantine, such as the *alif* and *tā' marbūṭa* variants in Arabic script and the definite-article vowel in the phonetic tier. ASR, G2P and A2P report the neutral tier, and the small grey figure beside each score gives the difference to the strict tier: the share of the error that is spelling or notation convention rather than recognition. Diacritization has its own fold (§3.1). The full rules are in Appendix B.1.

All dedicated systems are run zero-shot, while LLMs use a few-shot prompt derived from our annotation guidelines (with examples not from our benchmark). Full experimental details, including models (§B.4), configuration settings (§B.5), and prompt texts (§D), are provided in the Appendix.

### 3.1 Diacritization

**Task description.** Given an unvocalized Arabic utterance, predict the fully diacritized form, restoring the five diacritic categories mentioned in §2.3. The MSA case endings *-un*/*-in* are all but absent from the reference and, together with *sukun*, are normalized out of both references and predictions before scoring. The short vowels, *shadda*, and the accusative *tanwīn* are always scored.

| | Gemini | Opus 4.8 | GPT-5.5 | Shakkala | Mishkal | CATT | Levanti* |
|---|---|---|---|---|---|---|---|
| *DER (%)* | | | | | | | |
| **Overall** | **11.6** | 18.2 | 15.0 | 25.6 | 27.4 | 24.4 | 7.8 |
| Pal./Urban | **11.3** | 18.4 | 14.6 | 24.9 | 26.4 | 23.9 | 7.4 |
| Pal./Rural | **10.9** | 17.3 | 14.4 | 26.0 | 28.1 | 24.8 | 6.5 |
| Jordanian | **11.3** | 17.8 | 14.8 | 25.1 | 26.8 | 24.0 | 7.1 |
| Lebanese | **12.2** | 18.4 | 15.7 | 26.9 | 28.6 | 25.4 | 8.8 |
| Syrian | **12.1** | 18.8 | 15.6 | 25.1 | 27.2 | 24.2 | 9.3 |

Table 4: Diacritization results per variety (DER %, ↓) on all 1300 utterances. **Bold** = best per row, excluding Levanti (see asterisk). * Levanti produced silver diacritization labels used to annotate a portion of the dataset, so its score on those is not an independent measurement.

**Models.** We evaluate four dedicated systems; **Shakkala** (Barqawi, 2017), **Mishkal** (Zerrouki, 2014), **Catt** (Alasmary et al., 2024) (all designed for MSA), and **Levanti** (Mor-Lan, 2024) (designed for Levantine); alongside three LLMs: **Gemini 3.5 Flash**, **GPT 5.5**, and **Claude Opus 4.8**.

**Results.** Table 4 reports DER overall and by variety. Levanti reaches the lowest DER (7.8%), ahead of prompted Gemini 3.5 Flash (11.6%). GPT 5.5 (15.0%) and Claude Opus 4.8 (18.2%) form a middle tier, and the MSA-oriented Catt, Shakkala and Mishkal trail between 24.4% and 27.4%. Levanti's score is marked in the table, since it produced the silver diacritization used in annotating a portion of the utterances in the dataset (§2.3).

**Analysis.** Levantine adaptation matters more than system type or scale. The MSA-trained dedicated systems trail by a wide margin, while the dedicated Levantine system and the Gemini lead, with the larger LLMs forming a middle tier. A compact general LLM prompted with our Levantine-aware guidelines thus matches a purpose-built Levantine diacritizer, and outperforms both larger LLMs.

MSA systems predominantly over-mark word-final case endings. Our primary DER already excludes the MSA case-*tanwīn*, which occurs once in the reference; even so, these systems retain large residual error on the word-final vowels (*i'rāb*) that Levantine speech does not realize. Counting the case-*tanwīn* as well raises Catt, Shakkala and Mishkal to 26.8%, 27.9% and 28.8% without changing the ranking. Table 5 illustrates the pattern: MSA systems impose word-final case vowels and the definite article *al-* (vs. dialectal *il-*), whereas Levanti and Gemini track the dialectal reference.

| Reference | Model output |
|---|---|
| و كِيف بِدُّو يِعرِف اِنُّو خَبَّرنَاهُم<br>Jordanian | Catt: وَ **كَيْفَ** بَدُوٌّ **يُعْرَفُ** انْوِ خَبَرْنَاهُمْ |
| | Gemini: و كِيف بِدّو يِعرِف اِنّو خَبَّرنَاهُم |
| نِمشِي بَكِّير لَأَنُّه فِي هَالمَسَافِة كِلّهَا<br>Lebanese | Mishkal: نَمِشِيٌّ بِكِيرٍ لأَنّهُ فِي هالمسافة **كُلّهَا** |
| | Gemini: نِمشِي بَكِّير لَأَنُّه فِي هَالمَسَافِة كِلّهَا |
| نِحنَا هَلَّق بمَوسَم،<br>Lebanese | Mishkal: **نُحْنَا** هلق **بِمَوْسِمٍ** |
| | Levanti: نِحْنَا هَلَّق بْمَوْسِم، |

Table 5: Qualitative diacritization outputs: an MSA-oriented dedicated system (Catt, Mishkal) against a Levantine-aware system (Gemini, Levanti) on the same utterance. **Bold** marks words that still differ from the reference after normalization; the caption cites them undiacritized. Row 1 (Jordanian): Catt adds MSA case endings, a final vowel on كيف and يعرف that Levantine speech drops; Gemini matches the reference. Row 2 (Lebanese): Mishkal opens كلها with damma where the reference has kasra. Row 3 (Lebanese): Mishkal opens نحنا the same way, and vocalizes the prefix of بموسم, which Levantine leaves unvowelled; Levanti keeps the dialectal reading.

### 3.2 G2P

**Task description.** Given an undiacritized Arabic text, predict its phonetic transcription. The target uses the Semitic phonetic transcription system (§2.2). Because the systems emit different alphabets, outputs and targets are mapped onto the reference alphabet before scoring; the mapping is not exhaustive.

**Models.** We evaluate **CharsiuG2P** (Zhu et al., 2022) and **Levanti** (Mor-Lan, 2024) as dedicated G2P systems. For CharsiuG2P, we use the multilingual byT5 small model. We further test the LLMs Gemini 3.5 Flash, GPT-5.5, Claude Opus 4.8, and Jais v2 8B.

**Results.** Table 6 reports CER overall and by variety. The three frontier LLMs achieve similar CER, 11.9–12.2%, clearly outperforming the dedicated systems: Levanti (18.3%) and CharsiuG2P (34.8%).

**Analysis.** The gap between strict and neutral scoring reveals what drives each system's errors: Levanti's CER drops 4.1 points under neutral normalization, indicating much of its disadvantage is transcription convention (e.g., *āl-* article, *-eh* endings) rather than phonology; CharsiuG2P's gap is 2.4 points, confirming its errors are largely phonological—it restores MSA vocalization, gemination, and *q* for qaf. The LLMs show minimal strict–neutral differences ($\leq 0.7$), suggesting they already match the reference conventions. All systems struggle with rural Palestinian qaf, uniformly replacing the reference [k] with MSA [q] or urban [ʔ]. Table 7 illustrates these patterns. JAIS returned no transliteration for 131 of the 1,300 utterances, answering in Arabic script or replying to the utterance instead; those are scored as empty.

| | Gemini | Opus 4.8 | GPT-5.5 | JAIS-8B | Levanti* | Charsiu |
|---|---|---|---|---|---|---|
| *CER (%)* | | | | | | |
| **Overall** | **11.9** -0.4 | 12.0 -0.7 | 12.2 -0.4 | 40.8 -0.8 | 18.3 -4.1 | 34.8 -2.4 |
| Pal./Urban | 12.1 -0.3 | **11.0** -0.8 | 12.2 -0.3 | 41.8 -0.8 | 16.2 -4.0 | 32.3 -2.5 |
| Pal./Rural | **11.8** -1.2 | 11.9 -0.9 | 12.6 -1.2 | 43.0 -1.1 | 17.6 -3.7 | 36.1 -2.5 |
| Jordanian | 11.6 -0.2 | **11.3** -0.6 | 11.9 -0.3 | 36.6 -0.8 | 17.2 -3.7 | 34.0 -2.3 |
| Lebanese | **13.0** -0.1 | 13.6 -0.6 | 13.1 -0.1 | 39.1 -0.5 | 20.4 -4.0 | 36.5 -2.2 |
| Syrian | **11.2** -0.1 | 12.3 -0.6 | 11.3 -0.1 | 43.8 -0.8 | 20.2 -5.1 | 35.4 -2.6 |

Table 6: G2P results per variety (CER %, ↓) on all 1300 utterances, under *neutral* scoring. Small grey figures give the difference from the *strict* tier. **Bold** = best per row, excluding Levanti (see asterisk). * Levanti produced the expansion utterances' first-pass transcription (§2.3).

### 3.3 ASR

**Task description.** Given an audio recording, the model predicts the unvocalized Arabic text of the utterance, scored by WER and CER.

**Models.** We evaluate nine systems. Seven are dedicated speech recognizers: **SeamlessM4T-v2 large** (Barrault et al., 2023) in two configurations targeting MSA and Egyptian Arabic (**Seamless-MSA**, **Seamless-EGY**); **Whisper large v3** (Radford et al., 2023) and its two fine-tunes **Whisper-Levantine** and **BYZÖ-Whisper** (the latter covering eight Arabic dialects and serving as the third-ranking system at NADI 2025 (Talafha et al., 2025; M. Abdullah et al., 2025); it is fine-tuned on the *Casablanca* training split, so it is in-domain for the part of SHAMS drawn from that corpus, with no train/test overlap since we exclude the training splits); **MMS-1b** (Pratap et al., 2024); and **Cohere Transcribe** (Mack et al., 2026). Two are audio language models prompted to transcribe, **Gemini 3.5 Flash** and **GPT-Audio**. All are evaluated zero-shot with respect to SHAMS.

| Source | Output |
|---|---|
| *Notational — raw ↪ normalized* | |
| البصل مفروم<br>Lebanese | **Ref:** el-baṣal mafrūm ↪ ilbaṣal mafrūm<br>Charsiu: aːlbasˤal mafruːm<br>Levanti: ālbaṣal mafrūm<br>Claude: il-baṣal mafrūm<br>↪ *all normalize to* ilbaṣal mafrūm |
| *Phonological — persists after normalization* | |
| أنا من لبنان<br>Lebanese | **Ref:** ᵓana min libnēn<br>Charsiu: **ᵓannā minna lubnān**<br>Levanti: ᵓana min **lubnān**<br>Claude: ᵓana min **lubnān** |
| إلحق يا سيدي<br>Pal./Rural | **Ref:** ilḥak ya sīdi<br>Charsiu: **ilḥaqqa yā sēadī**<br>Levanti: **ᵓilḥaq** ya sīdi<br>Claude: **ilḥaᵓ** ya sīdi |

Table 7: G2P examples. *Top:* different raw notations converge after neutral normalization (↪). *Bottom* (normalized): phonological errors persist; **bold** marks words that differ from the normalized reference. CharsiuG2P adds MSA gemination/epenthesis; no system recovers the rural Palestinian qaf (*ḥak*, [k]), which is rendered as MSA [q] or urban [ᵓ].

**Results.** Table 8 reports WER and CER overall and by variety, on all 1,300 utterances. The two prompted audio LLMs and the dialect-targeted systems lead, with Gemini 3.5 Flash strongest overall by a clear margin (27.6 WER, 9.6 CER). Among the open-weight systems the separating factor is dialectal adaptation rather than architecture or scale: Seamless-EGY improves on its MSA-targeted counterpart on every variety, and BYZÖ-Whisper reaches the second-best CER overall.

**Analysis.** Models designed for MSA (Seamless-MSA, Whisper v3, MMS-1b) commonly output MSA equivalents of reference dialectal forms (Table 9), likely reflecting training on dialectal audio with MSA captions. Dialect-adapted systems are more faithful, including Seamless-EGY, likely because Egyptian and Levantine share lexical and grammatical features that diverge from MSA. The two scoring levels are most consequential for Whisper-Levantine, whose 15.0-point strict–neutral gap is by far the largest in the table: the model emits no أ, إ, آ, ة or ى, accounting for its large strict–neutral gap.

### 3.4 A2P

**Task description.** Given an audio recording, predict its phonetic transcription directly.

**Models.** We evaluate two audio-LLMs prompted with audio and textual instructions requesting a phonetic transcription: **GPT-Audio** and **Gemini 3.5 Flash**.

**Results.** Table 10 reports CER overall and by variety. Gemini 3.5 Flash attains the lowest error (20.5%), with GPT-Audio behind at 24.9%.

**Analysis.** Table 11 shows the two systems are usually faithful, differing mainly in notation and vowel quality; the dominant residual error is occasional reversion to Arabic script on hard utterances, which is scored as fully wrong.

We additionally evaluate both audio LLMs on ASR, which permits a within-model comparison of orthographic and phonetic transcription. Gemini 3.5 Flash reaches 9.6% CER on ASR against 20.5% on direct A2P, and GPT-Audio 14.4% against 24.9%, both tasks under the neutral tier. The two references, Arabic orthography and Semitic phonetic transcription, are not on a common scale, so the comparison is qualitative: both models transcribe orthography more reliably than phonetics.

To locate where the error arises, we compare direct A2P against an ASR → G2P cascade over the same utterances, holding the cascade's G2P stage fixed at Gemini so that the contrast is cascade versus direct rather than one cascade against another. The cascade reaches 18.9% CER for Gemini and 23.3% for GPT-Audio, against 20.5% and 24.9% for direct A2P, while a gold-transcript → G2P oracle reaches 13.0%. Direct phonetic transcription therefore offers no advantage over transcribing and then converting. Roughly 13 CER points are intrinsic to the phonetic conversion itself, with ASR error propagation adding a further six. Because the G2P stage is Gemini's in both cascades, the GPT-Audio cascade is a hybrid, and its margin over GPT-Audio's direct A2P is an upper bound. Per-variety figures for all three paths are in Table 14.

## 4 Related Work

**Diacritization and G2P.** Most Arabic diacritization work targets MSA or Classical Arabic (Zerrouki and Balla, 2017; Fadel et al., 2019; Alasmary et al., 2024; Mohamed and Mubarak, 2025), with recent character-level and pretrained approaches achieving strong results (Elmadany et al., 2022; Elgamal et al., 2024). Extension to dialects remains limited: Harrat et al. (2013) is an early effort on Algerian, Masmoudi et al. (2019) addresses Tunisian, while Shatnawi et al. (2024) and Ghannam et al. (2025) explore speech-informed and multimodal diacritic restoration. Arabic G2P has traditionally been decomposed into diacritization followed by rule-based phoneme mapping (Biadsy et al., 2009; Ali et al., 2014; Lodagala et al., 2025), while recently, Hawasly et al. (2025) introduced end-to-end G2P for Egyptian Arabic. A parallel gap for Modern Hebrew is addressed by Kolani et al. (2025) and Melichov et al. (2026), who introduce audio-grounded G2P benchmarks, enhanced diacritization, and speech-supervised G2P methods. Recent work has shown that off-the-shelf LLMs can outperform specialized Arabic diacritization models, though evaluations have focused exclusively on MSA (Toyin et al., 2025). To the best of our knowledge, no dedicated diacritization or G2P benchmark exists for LA.

| | Sml-MSA | Sml-EGY | Whisper v3 | Whisper-Lev. | MMS-1b | BYZÖ-W. | Cohere | Gemini | GPT Audio |
|---|---|---|---|---|---|---|---|---|---|
| *WER (%)* | | | | | | | | | |
| **Overall** | 40.4 -4.6 | 36.3 -4.1 | 39.3 -5.3 | 38.9 -15.0 | 72.5 -1.7 | 37.8 -6.0 | 35.5 -5.7 | **27.6** -5.4 | 34.9 -5.5 |
| Pal./Urban | 31.9 -4.9 | 29.1 -4.5 | 29.5 -6.5 | 35.3 -16.0 | 67.5 -2.1 | 39.2 -9.1 | 35.8 -5.7 | **26.5** -5.5 | 35.9 -5.5 |
| Pal./Rural | 47.6 -3.8 | 42.9 -2.7 | 45.4 -3.2 | 34.9 -14.2 | 77.9 -0.7 | **20.3** -3.1 | 33.2 -4.4 | 32.0 -2.9 | 41.2 -3.1 |
| Jordanian | 34.7 -5.8 | 31.8 -4.3 | 39.5 -5.5 | 30.3 -16.6 | 68.2 -1.9 | 34.4 -8.1 | 27.1 -6.5 | **21.0** -5.5 | 28.7 -5.5 |
| Lebanese | 44.3 -4.3 | 38.8 -4.5 | 45.7 -5.5 | 44.7 -14.1 | 76.3 -1.9 | 50.2 -4.9 | 42.5 -5.6 | **27.2** -6.1 | 34.0 -6.3 |
| Syrian | 43.9 -4.3 | 39.3 -4.5 | 36.9 -5.9 | 50.4 -14.4 | 72.9 -2.0 | 45.8 -4.7 | 39.6 -6.2 | **31.5** -7.2 | 34.5 -7.3 |
| *CER (%)* | | | | | | | | | |
| **Overall** | 17.8 -1.4 | 14.1 -1.2 | 20.6 -1.5 | 13.7 -3.9 | 27.0 -1.1 | 11.9 -1.6 | 16.8 -1.6 | **9.6** -1.5 | 14.4 -1.6 |
| Pal./Urban | 13.3 -1.5 | 10.8 -1.3 | 10.9 -1.8 | 11.4 -4.5 | 23.8 -1.2 | 11.4 -2.3 | 19.2 -1.6 | **9.7** -1.5 | 16.2 -1.7 |
| Pal./Rural | 19.1 -1.2 | 16.3 -0.8 | 32.0 -1.0 | 10.5 -3.8 | 29.3 -0.7 | **5.9** -0.8 | 11.7 -1.3 | 10.3 -0.9 | 16.6 -1.3 |
| Jordanian | 16.6 -1.6 | 12.8 -1.2 | 18.2 -1.5 | 10.1 -4.2 | 26.1 -1.1 | 10.3 -2.1 | 12.0 -1.8 | **7.1** -1.5 | 11.2 -1.6 |
| Lebanese | 19.3 -1.3 | 14.9 -1.2 | 24.0 -1.5 | 17.5 -3.3 | 29.0 -1.0 | 16.3 -1.5 | 22.8 -1.4 | **10.1** -1.4 | 14.7 -1.6 |
| Syrian | 21.0 -1.5 | 16.2 -1.4 | 18.9 -1.8 | 18.9 -3.9 | 27.2 -1.4 | 15.5 -1.3 | 18.2 -1.8 | **10.9** -2.1 | 13.2 -2.1 |

Table 8: ASR results per variety (WER/CER %, ↓) on all 1300 utterances, under *neutral* scoring. Small grey figures give the difference from the *strict* tier. **Bold** = best per row.

| Reference | MSA model | Dialect model |
|---|---|---|
| بس هيدي كانت مدرسة كتير منيحة<br>Lebanese | **لكن هذه** كانت مدرسة **جيدة جداً**<br>Seamless-MSA | بس **هادي** كانت مدرسة كتير منيحة<br>Seamless-EGY |
| لا مش بس أبو الفوز<br>Pal./Rural | **هذا ليس فقط** الفوز<br>Seamless-MSA | لا مش بس ابو الفوز<br>Whisper-Lev |
| ...بيطلع النوعين دراق بكفيا ودراق نبع الصفا<br>Lebanese | **...هناك نوعان من الدراجات: الدراجات الكافية و الدراجات الصالحة للسفر**<br>Seamless-MSA | |
| بدك إياني أقطع رزقي بإيدي؟<br>Jordanian | **هل تريد أن أضع** رزقي **في يديك**؟<br>Whisper-v3 | |

Table 9: MSA substitution in ASR. **Bold** marks words that differ from the reference. Rows 1–2 show all three: reference, MSA-biased output, and dialect-adapted output; the dialect model preserves the spoken forms. Rows 3–4 show further MSA errors (hallucinated vocabulary, full-sentence replacement).

| | Gemini | GPT-Audio |
|---|---|---|
| *CER (%)* | | |
| **Overall** | **20.5** -0.7 | 24.9 -0.5 |
| Pal./Urban | **21.1** -0.9 | 27.2 -0.6 |
| Pal./Rural | **21.0** -0.6 | 28.6 -0.5 |
| Jordanian | **18.0** -0.6 | 23.4 -0.4 |
| Lebanese | **20.8** -0.7 | 23.2 -0.5 |
| Syrian | **22.4** -0.7 | 23.4 -0.5 |

Table 10: A2P results per variety (CER %, ↓) on all 1300 utterances, under *neutral* scoring. Small grey figures give the difference from the *strict* tier. **Bold** = best per row.

**Dialectal Arabic Resources and ASR.** Dialectal corpora have grown from single-region datasets (Jarrar et al., 2017; Kwaik et al., 2018; Nayouf et al., 2023) to multi-dialectal collections (Bouamor et al., 2018; Talafha et al., 2024). For spoken Arabic, large resources such as Common Voice (Ardila et al., 2020) and MASC (Al-Fetyani et al., 2023) provide broad but coarse-grained coverage, often collapsing dialect distinc-tions into a single "Arabic" label (Lau et al., 2025). Dialect-specific ASR remains data-limited (Waheed et al., 2023; Djanibekov et al., 2025). The recent Arab Voices framework of Sullivan et al. (2026) spans dozens of datasets and dialects, but treats Levantine as a single variety and targets ASR alone. The NADI 2025 shared task (Talafha et al., 2025) recently introduced diacritic restoration for spoken dialects, but covers eight country-level dialects in a time-limited evaluation setting. Its Subtask 2 covered multidialectal ASR including Levantine, with six participating teams, notably Munsit (Salhab et al., 2025) and ELYADATA & LIA (Elleuch et al., 2025), along with related fine-tuning work such as M. Abdullah et al. (2025). AL-QASIDA (Robinson et al., 2025) evaluates LLM proficiency across eight dialectal varieties but focuses on text modeling rather than speech tasks. Cross-dialectal evaluation frameworks have been developed for other languages, notably Multi-VALUE for English (Ziems et al., 2023); no comparable resource exists for LA.

| Reference | GPT-Audio | Gemini 3.5 F. |
|---|---|---|
| šwayyit zēt<br>Lebanese | šwayyit zēt | šwayyit zēt |
| ᵓinti lāzim tistaġil-li hīk mōqif mitl ᶜā-datek<br>Jordanian | **inti** lāzim **tistaghillī hēk mawqif mitil ᶜādtik.** | **inti** lāzim **tistagallī hēk mawqif** mitl **ᶜādtik** |
| kamān il-ḥubb yaᶜni aḥyānan<br>Syrian | kamān il-ḥubb yaᶜni aḥyānan… | kamān il-ḥubb **yaᶜnī ᵓaḥyānan** |
| masalan, f-lnaftariḍ ᵓanno hīyye ᶜayy-nat ᶜala tlāt bnāt. …<br>Syrian | **بس أنا بفكر إنو هي عيلة على كلمة** | **matalan falan-naftariḍ innu hiyyi ᶜanīdi** ᶜala **talat banāt,** … |

Table 11: Qualitative A2P outputs. Words diverging from the reference are shown in **bold**. Row 1 (Lebanese) is transcribed perfectly by both systems. Row 2 (Jordanian) illustrates the typical case: minor vowel-quality drift (*hīk* → *hēk*), notation variants (*ġ* → *gh*), and reduced vowels. Row 3 (Syrian) is again near-perfect, with only long-vowel marking differences in Gemini. Row 4 exposes the dominant failure mode on a harder Syrian clip: GPT-Audio reverts entirely to Arabic script (scored as fully wrong), while Gemini stays in phonetic transcription but introduces errors throughout.

**Phone Recognition and Dialectal Transcription.** Transcribing speech directly to phones or phonemes has been studied under several names: multilingual phone recognition (Li et al., 2020; Zhu et al., 2025; Li et al., 2025), speech-to-IPA transcription for language documentation (Taguchi et al., 2023), audio-to-IPA models used to train and evaluate downstream speech systems (Kolani et al., 2025; Melichov et al., 2026), and automatic dialectal transcription for Finnish and Norwegian (Kuparinen, 2025).

## 5 Conclusion

We have presented SHAMS, a benchmark for Levantine Arabic speech-language technology comprising 1,300 utterances with four aligned annotation tiers. Our experiments on diacritization, G2P, ASR and A2P show that the benchmark effectively differentiates model performance across tasks and varieties, revealing both the progress of dialect-adapted models and LLMs and the substantial gaps that remain. The fine-grained annotations in SHAMS provide particular value in surfacing performance variation across different Levantine varieties, underscoring the importance of variety-aware evaluation.

## Limitations

The benchmark's sample size and coverage of the Levantine dialect space are both limited; for example, lacking Bedouin varieties, rural Jordanian, Lebanese, and Syrian varieties, and diaspora speech. Most varieties are drawn from a single source; for these, differences in content, speaker, and recording quality are confounded with dialect. Only Jordanian draws substantially on both and supports the within-variety source comparison (§B.2); there, the two sources also differ in speaker gender, so source and gender are entangled. Speaker gender is imbalanced in some varieties, which future work could mitigate. The Casablanca television data features actors who may not be native speakers of the portrayed variety, which could affect the representativeness of those segments. Inter-annotator agreement is reported for the diacritization tier over a 150-utterance subset (§2.4); the phonetic and orthographic tiers were each produced by a single expert. The set of mod-els and tasks tested could also be expanded, for example with audio-guided diacritization.

## Ethical Considerations

Our data is drawn from existing open audio corpora; licensing and release details are in Section 2.5. Annotations were produced by co-authors and collaborators with linguistics training, including native speakers of Palestinian Arabic (§2.3). This work is motivated by concerns about the under-representation of Arabic dialectal varieties in speech technology (Traberg et al., 2026; Fleisig et al., 2024; Cerqueglini, 2026); our benchmark provides an instrument for measuring how well current systems serve LA speakers across varieties, and supporting work on increasing this representation.

## Acknowledgements

We thank Bessan Adi and Ivan Khuri for their contributions to the annotation effort, and Shai Gordin for providing infrastructure enabling this project.

## References

Mohammad Al-Fetyani, Muhammad Al-Barham, Gheith Abandah, Adham Alsharkawi, and Maha Dawas. 2023. Masc: Massive arabic speech corpus. In *2022 IEEE Spoken Language Technology Workshop (SLT)*, pages 1006–1013. IEEE.

Enam Al-Wer. 2007. The formation of the dialect of amman: From chaos to order. In *Arabic in the City*, pages 69–90. Routledge.

Enam Al-Wer, Uri Horesh, Bruno Herin, and Rudolf De Jong. 2022. *Arabic sociolinguistics*. Cambridge University Press.

Faris Alasmary, Orjuwan Zaafarani, and Ahmad Ghannam. 2024. Catt: Character-based arabic tashkeel transformer. In *Proceedings of the Second Arabic Natural Language Processing Conference*, pages 250–257.

Ahmed Ali, Yifan Zhang, Patrick Cardinal, Najim Dahak, Stephan Vogel, and James Glass. 2014. A complete kaldi recipe for building arabic speech recognition systems. In *2014 IEEE spoken language technology workshop (SLT)*, pages 525–529. IEEE.

Rosana Ardila, Megan Branson, Kelly Davis, Michael Kohler, Josh Meyer, Michael Henretty, Reuben Morais, Lindsay Saunders, Francis Tyers, and Gregor Weber. 2020. Common voice: A massively-multilingual speech corpus. In *Proceedings of the twelfth language resources and evaluation conference*, pages 4218–4222.

Ahmad Barqawi. 2017. Shakkala: Arabic text vocalization using deep learning. `https://github.com/Barqawiz/Shakkala`.

Loïc Barrault, Yu-An Chung, Mariano Cora Meglioli, David Dale, Ning Dong, Paul-Ambroise Duquenne, Hady Elsahar, Hongyu Gong, Kevin Heffernan, John Hoffman, and 1 others. 2023. Seamlessm4t: Massively multilingual & multimodal machine translation. *arXiv preprint arXiv:2308.11596*.

Fadi Biadsy, Nizar Habash, and Julia Hirschberg. 2009. Improving the arabic pronunciation dictionary for phone and word recognition with linguistically-based pronunciation rules. In *Proceedings of human language technologies: The 2009 annual conference of the North American chapter of the association for computational linguistics*, pages 397–405.

Houda Bouamor, Nizar Habash, Mohammad Salameh, Wajdi Zaghouani, Owen Rambow, Dana Abdulrahim, Ossama Obeid, Salam Khalifa, Fadhl Eryani, Alexander Erdmann, and 1 others. 2018. The madar arabic dialect corpus and lexicon. In *Proceedings of the eleventh international conference on language resources and evaluation (LREC 2018)*.

Kristen Brustad and Emilie Zuniga. 2019. Levantine arabic. In *The Semitic Languages*, pages 403–432. Routledge.

L. Cerqueglini. 2026. Humic (human–machine interaction), university of surrey, guildford, june 16th 2026. In *Lēš byiḥčīš? ('Why Don't You Speak?'): Conversational AI at the Margins – The Case of Palestinian Arabic*.

Letizia Cerqueglini. 2022. Negev bedouin daḥīyah songs: Celebrating brotherhood above war and love. *Folia Orientalia*, 59:95–126.

Letizia Cerqueglini. 2024. Ya ṛaḅḅí ! tanṭīni biné:t!: Ǧbēne in the traditional muṯallaṯ arabic of bāḳa l-ġarbiyye. *Folia Orientalia*, 60:227–275.

Peter T. Daniels. 1990. Fundamentals of grammatology. *Journal of the American Oriental Society*, 110(4):727–731.

Amirbek Djanibekov, Hawau Olamide Toyin, Raghad Alshalan, Abdullah Alatir, and Hanan Aldarmaki. 2025. Dialectal coverage and generalization in arabic speech recognition. In *Proceedings of the 63rd Annual Meeting of the Association for Computational Linguistics (Volume 1: Long Papers)*, pages 29490–29502.

Salman Elgamal, Ossama Obeid, Mhd Kabbani, Go Inoue, and Nizar Habash. 2024. Arabic diacritics in the wild: Exploiting opportunities for improved diacritization. In *Proceedings of the 62nd Annual Meeting of the Association for Computational Linguistics (Volume 1: Long Papers)*, pages 14815–14829.

Haroun Elleuch, Youssef Saidi, Salima Mdhaffar, Yannick Estève, and Fethi Bougares. 2025. ELYADATA

& LIA at NADI 2025: ASR and ADI subtasks. In *Proceedings of The Third Arabic Natural Language Processing Conference: Shared Tasks*, pages 762–766, Suzhou, China. Association for Computational Linguistics.

AbdelRahim Elmadany, Muhammad Abdul-Mageed, and 1 others. 2022. Arat5: Text-to-text transformers for arabic language generation. In *Proceedings of the 60th annual meeting of the association for computational linguistics (Volume 1: Long papers)*, pages 628–647.

Ali Fadel, Ibraheem Tuffaha, Mahmoud Al-Ayyoub, and 1 others. 2019. Neural arabic text diacritization: State of the art results and a novel approach for machine translation. In *Proceedings of the 6th Workshop on Asian Translation*, pages 215–225.

Charles A Ferguson. 1959. Diglossia. *word*, 15(2):325–340.

Five College Center for World Languages. n.d. LangMedia: Arabic. https://www.langmedia.fivecolleges.edu/languages/arabic. Accessed 2026-06-15.

Eve Fleisig, Genevieve Smith, Madeline Bossi, Ishita Rustagi, Xavier Yin, and Dan Klein. 2024. Linguistic bias in ChatGPT: Language models reinforce dialect discrimination. In *Proceedings of the 2024 Conference on Empirical Methods in Natural Language Processing*, pages 13541–13564, Miami, Florida, USA. Association for Computational Linguistics.

Ahmad Ghannam, Naif Alharthi, Faris Alasmary, Kholood Al Tabash, Shouq Sadah, and Lahouari Ghouti. 2025. Abjad ai at nadi 2025: Catt-whisper: Multimodal diacritic restoration using text and speech representations. In *Proceedings of The Third Arabic Natural Language Processing Conference: Shared Tasks*, pages 757–761.

Nizar Y Habash. 2010. *Introduction to Arabic natural language processing*. Morgan & Claypool Publishers.

Salima Harrat, Mourad Abbas, Karima Meftouh, and Kamel Smaili. 2013. Diacritics restoration for arabic dialects. In *INTERSPEECH 2013-14th Annual Conference of the International Speech Communication Association*.

Majd Hawasly, Hamdy Mubarak, Ahmed Abdelali, and Ahmed Ali. 2025. Dialg2p: Dialectal grapheme-to-phoneme. arabic as a case study. In *Proceedings of The Third Arabic Natural Language Processing Conference*, pages 466–471.

Roni Henkin. 2010. *Negev Arabic: Dialectal, sociolinguistic, and stylistic variation*. Harrassowitz Verlag.

Mustafa Jarrar, Nizar Habash, Faeq Alrimawi, Diyam Akra, and Nasser Zalmout. 2017. Curras: an annotated corpus for the palestinian arabic dialect. *Language Resources and Evaluation*, 51(3):745–775.

Yakov Kolani, Maxim Melichov, Cobi Calev, and Morris Alper. 2025. Phonikud: Hebrew grapheme-to-phoneme conversion for real-time text-to-speech. *arXiv preprint arXiv:2506.12311*.

Olli Kuparinen. 2025. Automatic dialectal transcription: An evaluation on finnish and norwegian. In *Proc. Interspeech 2025*, pages 2390–2394.

Kathrein Abu Kwaik, Motaz Saad, Stergios Chatzikyriakidis, and Simon Dobnik. 2018. Shami: A corpus of levantine arabic dialects. In *Proceedings of the eleventh international conference on language resources and evaluation (LREC 2018)*.

Mingfei Lau, Qian Chen, Yeming Fang, Tingting Xu, Tongzhou Chen, and Pavel Golik. 2025. Data quality issues in multilingual speech datasets: The need for sociolinguistic awareness and proactive language planning. In *Proceedings of the 63rd Annual Meeting of the Association for Computational Linguistics (Volume 1: Long Papers)*, pages 7466–7492.

Chin-Jou Li, Kalvin Chang, Shikhar Bharadwaj, Eunjung Yeo, Kwanghee Choi, Jian Zhu, David Mortensen, and Shinji Watanabe. 2025. Powsm: A phonetic open whisper-style speech foundation model. *arXiv preprint arXiv:2510.24992*.

Xinjian Li, Siddharth Dalmia, Juncheng Li, Matthew Lee, Patrick Littell, Jiali Yao, Antonios Anastasopoulos, David R Mortensen, Graham Neubig, Alan W Black, and 1 others. 2020. Universal phone recognition with a multilingual allophone system. In *ICASSP 2020-2020 IEEE International Conference on Acoustics, Speech and Signal Processing (ICASSP)*, pages 8249–8253. IEEE.

Vasista Sai Lodagala, Lamya Alkanhal, Daniel Izham, Shivam Mehta, Shammur Absar Chowdhury, Aqeelah Makki, Hamdy S Hussein, Gustav Eje Henter, and Ahmed Ali. 2025. Sawtarabi: A benchmark corpus for arabic tts. standard, dialectal and code-switching. In *Proc. Interspeech 2025*, pages 4793–4797.

Badr M. Abdullah, Yusser Al Ghussin, Zena Al-Khalili, Ömer Tarik Özyilmaz, Matias Valdenegro-Toro, Simon Ostermann, and Dietrich Klakow. 2025. Saarland-Groningen at NADI 2025 shared task: Effective dialectal Arabic speech processing under data constraints. In *Proceedings of The Third Arabic Natural Language Processing Conference: Shared Tasks*, pages 745–751, Suzhou, China. Association for Computational Linguistics.

Julian Mack, Ekagra Ranjan, Walter Beller-Morales, Bharat Venkitesh, and Pierre Richemond. 2026. cohere-transcribe-03-2026 (revision d96e814).

Abir Masmoudi, Salima Mdhaffar, Rahma Sellami, and Lamia Hadrich Belguith. 2019. Automatic diacritics restoration for tunisian dialect. *ACM Transactions on Asian and Low-Resource Language Information Processing (TALLIP)*, 18(3):1–18.

Maxim Melichov, Yakov Kolani, and Morris Alper. 2026. Renikud: Audio-supervised hebrew grapheme-to-phoneme conversion. *arXiv preprint arXiv:2606.20179*.

Abubakr Mohamed and Hamdy Mubarak. 2025. Advancing arabic diacritization: Improved datasets, benchmarking, and state-of-the-art models. In *Proceedings of the 2025 Conference on Empirical Methods in Natural Language Processing*, pages 16707–16719.

Guy Mor-Lan. 2024. Levanti: A levantine arabic dataset. https://huggingface.co/datasets/guymorlan/levanti.

Amal Nayouf, Tymaa Hammouda, Mustafa Jarrar, Fadi Zaraket, and Mohamad-Bassam Kurdy. 2023. Nâbra: Syrian arabic dialects with morphological annotations. In *Proceedings of ArabicNLP 2023*, pages 12–23.

Vineel Pratap, Andros Tjandra, Bowen Shi, Paden Tomasello, Arun Babu, Sayani Kundu, Ali Elkahky, Zhaoheng Ni, Apoorv Vyas, Maryam Fazel-Zarandi, and 1 others. 2024. Scaling speech technology to 1,000+ languages. *Journal of Machine Learning Research*, 25(97):1–52.

Alec Radford, Jong Wook Kim, Tao Xu, Greg Brockman, Christine McLeavey, and Ilya Sutskever. 2023. Robust speech recognition via large-scale weak supervision. In *International conference on machine learning*, pages 28492–28518. PMLR.

Nathaniel Romney Robinson, Shahd Abdelmoneim, Kelly Marchisio, and Sebastian Ruder. 2025. Alqasida: Analyzing llm quality and accuracy systematically in dialectal arabic. In *Findings of the Association for Computational Linguistics: ACL 2025*, pages 22048–22065.

Mahmoud Salhab, Shameed Sait, Mohammad Abusheikh, and Hasan Abusheikh. 2025. Munsit at NADI 2025 shared task 2: Pushing the boundaries of multidialectal Arabic ASR with weakly supervised pretraining and continual supervised fine-tuning. In *Proceedings of The Third Arabic Natural Language Processing Conference: Shared Tasks*, pages 734–739, Suzhou, China. Association for Computational Linguistics.

Kimary N. Shahin. 2011. Palestinian arabic. Brill, Leiden, The Netherlands.

Sara Shatnawi, Sawsan Alqahtani, and Hanan Aldarmaki. 2024. Automatic restoration of diacritics for speech data sets. In *Proceedings of the 2024 Conference of the North American Chapter of the Association for Computational Linguistics: Human Language Technologies (Volume 1: Long Papers)*, pages 4166–4176.

Peter Sullivan, AbdelRahim Elmadany, Alcides Alcoba Inciarte, and Muhammad Abdul-Mageed. 2026. Arab voices: Mapping standard and dialectal arabic speech technology. *arXiv preprint arXiv:2601.13319*.

Chihiro Taguchi, Yusuke Sakai, Parisa Haghani, and David Chiang. 2023. Universal automatic phonetic transcription into the international phonetic alphabet. *arXiv preprint arXiv:2308.03917*.

Bashar Talafha, Karima Kadaoui, Samar Mohamed Magdy, Mariem Habiboullah, Chafei Mohamed Chafei, Ahmed Oumar El-Shangiti, Hiba Zayed, Mohamedou Cheikh Tourad, Rahaf Alhamouri, Rwaa Assi, and 1 others. 2024. Casablanca: Data and models for multidialectal arabic speech recognition. In *Proceedings of the 2024 Conference on Empirical Methods in Natural Language Processing*, pages 21745–21758.

Bashar Talafha, Hawau Olamide Toyin, Peter Sullivan, AbdelRahim A Elmadany, Abdurrahman Juma, Amirbek Djanibekov, Chiyu Zhang, Hamad Alshehhi, Hanan Aldarmaki, Mustafa Jarrar, and 1 others. 2025. Nadi 2025: The first multidialectal arabic speech processing shared task. In *Proceedings of The Third Arabic Natural Language Processing Conference: Shared Tasks*, pages 720–733.

Hawau Olamide Toyin, Samar M Magdy, and Hanan Aldarmaki. 2025. Are llms good text diacritizers? an arabic and yor\ub\'a case study. *arXiv preprint arXiv:2506.11602*.

Cecilie Steenbuch Traberg, Jon Roozenbeek, and Sander Van Der Linden. 2026. Ai is turning research into a scientific monoculture. *Communications Psychology*, 4(1):37.

Kees Versteegh, editor. 2006–2009. *Encyclopedia of Arabic Language and Linguistics*, volume 1–5. Brill, Leiden.

Abdul Waheed, Bashar Talafha, Peter Sullivan, AbdelRahim Elmadany, and Muhammad Abdul-Mageed. 2023. Voxarabica: A robust dialect-aware arabic speech recognition system. In *Proceedings of ArabicNLP 2023*, pages 441–449.

Taha Zerrouki. 2014. Mishkal: Arabic text vocalization software. https://github.com/linuxscout/mishkal.

Taha Zerrouki and Amar Balla. 2017. Tashkeela: Novel corpus of arabic vocalized texts, data for autodiacritization systems. *Data in brief*, 11:147.

Jian Zhu, Farhan Samir, Eleanor Chodroff, and David R Mortensen. 2025. Zipa: A family of efficient models for multilingual phone recognition. In *Proceedings of the 63rd Annual Meeting of the Association for Computational Linguistics (Volume 1: Long Papers)*, pages 19568–19585.

Jian Zhu, Cong Zhang, and David Jurgens. 2022. Byt5 model for massively multilingual grapheme-to-phoneme conversion.

Caleb Ziems, William Held, Jingfeng Yang, Jwala Dhamala, Rahul Gupta, and Diyi Yang. 2023. Multivalue: A framework for cross-dialectal english nlp.

In *Proceedings of the 61st Annual Meeting of the Association for Computational Linguistics (Volume 1: Long Papers)*, pages 744–768.

## A Further Annotation Details

This appendix provides detailed annotation conventions used in SHAMS.

**Diacritization principles.** Our annotation scheme balances phonetic fidelity with preservation of the original orthographic form, guided by readability, consistency, and recoverability of the underlying word. Diacritics capture dialectal pronunciation contrasts not reflected in the consonantal skeleton—for example, distinguishing عَلِي from عَلَيّ, or صُوت from صَوت—without introducing alternative spellings.

Original consonantal forms are generally preserved. The letter ق is retained regardless of whether it is realized as /q/, a glottal stop, /g/, or another variant; *imāla* is represented through vowel diacritics rather than changes to the underlying letters (e.g., سِاعَة rather than سيعة). Consonantal substitutions are reflected only when they do not create ambiguity and improve representation of dialectal speech.

Gemination is consistently marked with *shadda*, as it may affect lexical meaning and morphological structure. Assimilated or reduced forms are segmented into their underlying components (e.g., مِن هُن rather than مِنُّن) to preserve morphological transparency. Dialect-specific lexical and morphological features such as هيدي and verbal patterns such as مِنروح versus بِنروح are retained rather than normalized, as they provide valuable cues for dialect identification.

**Transcription conventions.** Word-initial *alif* ranges from a full glottal stop to complete elision depending on prosodic context. For example, in 46392_1: *haḏōl dār Abu Fījli. yā salām! ʾawwal ʾiši ...*—*ʾawwal* and *ʾiši* have glottal onset, whereas *Abu* does not.

Integrated loanwords are transcribed as pronounced rather than according to their source-language spelling (e.g., 46857_1: French *colonel* and *général* are realized as *il-kulunēl il-ǧinirāl*).

**Phonetic variation.** Because the tier records the realization in the recording, one orthographic word can carry several phonetic forms across the corpus. The reflex of ق varies: قاعد ('sitting') appears as *qāʿid*, *gāʿid*, *kāʿid* and *ʾāʿed*, and فوق ('above') as *fōq*, *fōʾ* and *fōk*. Affrication of ك appears in the Rural Palestinian material, where كل ('all') is transcribed *čill* beside *kull* and *kill*, and كفر, in the place name *Kafr il-Lōz*, as both *Kafr* and *Čafr*. The interdentals are kept, stopped or turned into sibilants: مثلا ('for example') gives *maṯalan*, *matalan* and *masalan*, and اذا ('if') gives *ʾiḏa* and *ʾiza*.

The vowels vary along several axes. *Imāla* raises the long *ā*: كمان ('also') appears as *kamān* and, in the Lebanese material, as *kamēn*, and لازم ('must') as *lāzim* and *lēzim*. Diphthongs either hold or collapse, so موسم ('season') appears as *mawsam* and *mōsam*, and صيف ('summer') as *ṣayf* and *ṣēf*. The long mid vowels raise, giving هون ('here') as *hōn* and *hūn*, and بيت ('house') as *bēt* and *bīt*. Short vowels are inserted or dropped, giving مثل ('like') as *mitl* and *mitil*.

Variation of this kind is not only between varieties. الفوز, part of the name *Abu l-Fōz*, splits between *il-Fawz* and *l-Fōz* within the Rural Palestinian material, and هيدا ('this') between *hayda* and *hēda* within the Lebanese.

## B Experimental Details

### B.1 Preprocessing and Normalization

The release already carries Unicode canonicalization at dataset construction (§2.1): the released tiers are canonically composed, with presentation forms folded, invisible controls and *tatweel* removed. We repeat the same step at scoring time, since system output is not released text. Everything below is an evaluation-time normalization, applied symmetrically to references and hypotheses. The fold is the reductive orthographic normalization of Habash (2010): the letter normalizations most Arabic NLP work applies by default. We do not fold the non-*alif* forms of *hamza*.

Both Arabic-script tiers, orthographic and diacritized, share a standard **orthographic normalization**: *alif* variants (أ, إ, آ, ٱ → ا), *alif maqṣūra* (ى → ي) and *tā' marbūṭa* (ة → ه). These distinctions carry no phonemic contrast in Levantine and systems spell them inconsistently. Normalizing them prevents spelling differences from counting as errors.

**Diacritization.** We apply this orthographic normalization, but keep all other diacritics since they are the evaluation target. Three steps are added. We strip *sukun*, since gold labels are sukun-free by construction. We canonicalize accusative *tanwin* onto the consonant preceding the seat alif, so that كتاباً and كتابًا score as identical. And we drop the MSA case-*tanwin*, *dammatan* and *kasratan*, from both sides while keeping *fathatan*, the realized *-an*: spoken Levantine does not realize MSA case endings, so a system that supplies them is displaying MSA register rather than making a diacritization error. These endings are almost absent from the reference, 1 of 15,506 tokens. §3.1 reports the *tanwin*-included DER as an ablation.

**ASR.** Both tiers strip punctuation and collapse whitespace. The *neutral* tier also applies the orthographic normalization above and strips residual diacritics; the *strict* tier omits both and scores the orthography as written. Reporting the two separates recognition error from spelling convention: the difference is large only for systems trained on normalized text, and is given beside each neutral figure in Table 8.

**G2P and A2P.** The phonetic tier is Latin-script, so the Arabic-script normalization does not apply. The *strict* tier maps every output onto the reference alphabet (§2.2) by character rules (e.g., /ɣ/→ġ, /ʃ/→š, /aː/→ā, *ey*→ē) and strips segmentation hyphens, punctuation and casing. The *neutral* tier, the primary tier for both tasks, additionally folds notation choices that not every system supports: the definite-article vowel and sun-letter assimilation (*āl-*/*il-*), the *tāʾ marbūṭa* ending (*-eh*/*-e*, *-ah*/*-a*), the pairs *ǧ*/*j* and *č*/*k*, and the reduced vowel *ə*. Phonological contrasts (vowel length and quality, emphatics, gemination, qaf realization) are kept under both tiers. Table 7 shows raw and normalized outputs. The mapping is not exhaustive: notation the rules do not cover is scored as error under both tiers. For A2P, each per-utterance CER is capped at 100% before averaging, to bound looping or runaway audio-LLM output.

## B.2 Per-Source Results

Two varieties draw on both corpora: Jordanian, 137 *Casablanca* utterances against 168 from *LangMedia* (Table 12), and Palestinian Urban, 22 against 208 (Table 13). For Palestinian Rural, Lebanese and Syrian the per-variety figures in Table 8 are the per-source figures, since each is drawn from one corpus. On Jordanian the direction of the source effect depends on the system. *LangMedia* is harder for six of the nine systems (from $+0.4$ to $+10.0$ WER); Whisper v3 and MMS-1b are better on it ($-26.3$ and $-14.5$), and GPT-Audio is level ($-1.3$). BYZÖ-Whisper, fine-tuned on the *Casablanca* training split, is better on *Casablanca* material in both varieties ($+7.5$ Jordanian, $+11.2$ Palestinian Urban) and overall (23.7 against 44.2 WER). Source sensitivity here tracks training exposure. The two Jordanian sources also differ in speaker gender (§5).

| | Casablanca $n = 137$, 10.5 min | LangMedia $n = 168$, 15.0 min | Δ |
|---|---|---|---|
| *WER (%)* | | | |
| Sml-MSA | 34.5 | 34.9 | +0.4 |
| Sml-EGY | 29.0 | 34.1 | +5.1 |
| Whisper v3 | 53.8 | 27.4 | -26.3 |
| Whisper-Lev. | 24.8 | 34.8 | +10.0 |
| MMS-1b | 76.1 | 61.6 | -14.5 |
| BYZÖ-W. | 30.3 | 37.8 | +7.5 |
| Cohere | 22.9 | 30.5 | +7.6 |
| Gemini | 19.4 | 22.2 | +2.8 |
| GPT Audio | 29.4 | 28.1 | -1.3 |
| *CER (%)* | | | |
| Sml-MSA | 15.1 | 17.9 | +2.8 |
| Sml-EGY | 8.8 | 16.0 | +7.2 |
| Whisper v3 | 27.7 | 10.7 | -17.0 |
| Whisper-Lev. | 7.4 | 12.1 | +4.7 |
| MMS-1b | 29.0 | 23.9 | -5.1 |
| BYZÖ-W. | 8.3 | 11.9 | +3.5 |
| Cohere | 7.9 | 15.3 | +7.4 |
| Gemini | 5.9 | 8.0 | +2.1 |
| GPT Audio | 11.5 | 11.0 | -0.5 |

Table 12: Within-variety source contrast for Jordanian, scored as in Table 8, the within-variety comparison the expanded dataset was selected to support. Δ is LangMedia − Casablanca, and is a difference of rates rather than an estimate with an interval; read it against the two utterance counts in the header. Source and speaker gender remain partly entangled (§5), so this bounds rather than isolates the source effect.

Each task's primary score is the one shown in its main table. Where a small grey number appears next to a score, it shows the difference from the alternative tier.

## B.3 Per-Variety Audio Results

Per-variety WER and CER for all nine ASR systems are given in the main text (Table 8). Table 14 gives the same breakdown for the A2P comparison in §3.4: direct A2P, the ASR → G2P cascade, and the gold-transcript → G2P oracle.

The per-variety numbers add two points to the overall ones. The oracle ranges over 3.5 points across varieties (11.4 to 14.9), against 4.5 and 5.4 for the two direct A2P systems, so most of the per-variety spread comes from the audio stage and not from G2P. And the cascade does not beat direct A2P everywhere: on Palestinian/Rural Gemini's cascade is worse than its direct A2P (22.4 against 21.0), and on Lebanese GPT-Audio's cascade margin is nil (23.1 against 23.2). Palestinian/Rural is also the oracle's hardest variety, so it is where the shared G2P stage is weakest.

| | Casablanca $n = 22$, 1.9 min | LangMedia $n = 208$, 23.1 min | Δ |
|---|---|---|---|
| *WER (%)* | | | |
| Sml-MSA | 39.8 | 31.2 | -8.6 |
| Sml-EGY | 36.5 | 28.4 | -8.0 |
| Whisper v3 | 29.3 | 29.6 | +0.2 |
| Whisper-Lev. | 32.0 | 35.6 | +3.7 |
| MMS-1b | 64.7 | 67.8 | +3.2 |
| BYZÖ-W. | 28.9 | 40.1 | +11.2 |
| Cohere | 31.6 | 36.2 | +4.6 |
| Gemini | 29.3 | 26.3 | -3.1 |
| GPT Audio | 39.5 | 35.6 | -3.9 |
| *CER (%)* | | | |
| Sml-MSA | 10.8 | 13.5 | +2.7 |
| Sml-EGY | 10.6 | 10.8 | +0.1 |
| Whisper v3 | 8.2 | 11.1 | +2.9 |
| Whisper-Lev. | 8.6 | 11.6 | +3.0 |
| MMS-1b | 23.6 | 23.8 | +0.3 |
| BYZÖ-W. | 6.3 | 11.8 | +5.5 |
| Cohere | 10.3 | 19.9 | +9.6 |
| Gemini | 8.2 | 9.8 | +1.6 |
| GPT Audio | 13.8 | 16.3 | +2.5 |

Table 13: Within-variety source contrast for Pal./Urban, scored as in Table 8, small on the Casablanca side and therefore indicative rather than conclusive. Δ is LangMedia − Casablanca, and is a difference of rates rather than an estimate with an interval; read it against the two utterance counts in the header. Source and speaker gender remain partly entangled (§5), so this bounds rather than isolates the source effect.

### B.4 Models

Table 15 lists the diacritization models evaluated in our experiments with links and configuration details, Table 16 the G2P models, and Table 17 the ASR, and Table 18 the A2P models.

### B.5 LLM Generation Settings

For reproducibility, Table 19 reports the decoding settings used for every prompted LLM. All prompted LLMs are accessed through their official APIs at the identifiers listed in Table 15 and Table 18, except for JAIS which is served locally with 4-bit quantization.

For the text-input tasks (diacritization and G2P) we request a JSON output and parse it with fallbacks. For A2P we request a single line of transliteration directly from the waveform (`gpt-audio` and `gemini-3.5-flash` accept audio natively). We cap the maximum number of output tokens to bound the rare runaway/looping generation.

### B.6 Example Outputs

Table 20, Table 21, Table 22, and Table 23 show full-utterance diacritization, G2P, ASR and A2P outputs for one example per variety, alongside the gold reference.

## C Generative AI Disclosure

Generative AI tools were used for polishing manuscript wording and coding assistance.

| Audio | Path | All | Pal./Urban | Pal./Rural | Jordanian | Lebanese | Syrian |
|---|---|---|---|---|---|---|---|
| Gemini 3.5 Flash | direct A2P | 20.5 | 21.1 | **21.0** | 18.0 | 20.8 | 22.4 |
| | ASR → G2P cascade | **18.9** | **18.7** | 22.4 | **16.3** | **18.6** | **19.3** |
| GPT-Audio | direct A2P | 24.9 | 27.2 | 28.6 | 23.4 | 23.2 | 23.4 |
| | ASR → G2P cascade | 23.3 | 24.6 | 27.5 | 21.0 | 23.1 | 20.7 |
| Gold transcript* | G2P oracle | 13.0 | 11.4 | 14.9 | 12.1 | 13.9 | 12.6 |

Table 14: Per-variety CER (%, ↓) for direct A2P against an ASR → G2P cascade and a gold-transcript → G2P oracle, on all 1300 utterances. The G2P stage is Gemini 3.5 Flash in every cascade row, so the GPT-Audio cascade is a hybrid and its margin over GPT-Audio's direct A2P is an upper bound. Scored with the G2P *neutral* normalization and the 100% per-utterance cap (§B.1); figures are macro means over utterances. **Bold** = best per column among the four audio systems. * Cascade: gold text is not an audio system. It passes the gold orthographic transcription through the same G2P module, and is the floor the two cascades are measured against rather than a system competing with them.

| Model | Config | Source |
|---|---|---|
| Shakkala | char-BiLSTM, zero-shot | Barqawi (2017) |
| Mishkal | rule-based, zero-shot | Zerrouki (2014) |
| Catt | Transformer, zero-shot | Alasmary et al. (2024) |
| Levanti* | Levantine, zero-shot | Mor-Lan (2024) |
| Gemini 3.5 Flash | few-shot prompt | `gemini-3.5-flash` |
| GPT 5.5 | few-shot prompt | `gpt-5.5` |
| Claude Opus 4.8 | few-shot prompt | `claude-opus-4-8` |

Table 15: Diacritization models evaluated. Dedicated systems are run zero-shot; the three prompted LLMs use the few-shot prompt in Appendix D.1. Shakkala, Mishkal, and Catt were developed for Modern Standard Arabic, whereas Levanti is trained on Levantine data. * Levanti produced the silver diacritization labels that the 780 expansion utterances were annotated from, so its score on those is not an independent measurement.

| System | Configuration |
|---|---|
| Gemini 3.5 Flash | prompted |
| Claude Opus 4.8 | prompted |
| GPT-5.5 | prompted |
| JAIS v2 8B | prompted |
| Levanti | — |
| CharsiuG2P | multilingual byT5 small |

Table 16: G2P systems evaluated, with configuration and model URLs.

| System | Configuration |
|---|---|
| Seamless-MSA | `tgt_lang=arb` |
| Seamless-Egyptian | `tgt_lang=arz` |
| Whisper v3 | — |
| Whisper-Levantine | — |
| MMS-1B | `ara` adapter |
| BYZÖ-Whisper | — |
| Cohere Transcribe | `language=ar` |
| Gemini 3.5 Flash | — |
| GPT-Audio | — |

Table 17: ASR systems evaluated, with configuration and model URLs.

| Model | Config | Source |
|---|---|---|
| GPT-Audio | direct, prompted, audio in | `gpt-audio` |
| Gemini 3.5 Flash | direct, prompted, audio in | `gemini-3.5-flash` |
| GPT-Audio ASR → G2P | cascade, Gemini 3.5 Flash G2P | `gpt-audio`, `gemini-3.5-flash` |
| Gemini ASR → G2P | cascade, Gemini 3.5 Flash G2P | `gemini-3.5-flash` |
| Gold text → G2P* | oracle, Gemini 3.5 Flash G2P | `gemini-3.5-flash` |

Table 18: Audio-to-phoneme (A2P) systems. The two direct systems are audio language models prompted to output a single-line phonetic transcription from the waveform, with the identical prompt. The cascades pass an ASR transcript through the Gemini 3.5 Flash G2P module. * The oracle passes the gold orthographic transcription through the same module and is not an audio system. Outputs are scored after the shared G2P normalization (§B.1).

| Task | Model | Serving | Precision | Temp. | Max tok. | Thinking |
|---|---|---|---|---|---|---|
| Diac. | Gemini 3.5 Flash | API | — | 0 | 4096 | minimal |
| Diac. | GPT 5.5 | API | — | default | 4096 | none |
| Diac. | Claude Opus 4.8 | API | — | default | 4096 | default |
| G2P | Gemini 3.5 Flash | API | — | default | default | default |
| G2P | GPT 5.5 | API | — | default | default | default |
| G2P | Claude Opus 4.8 | API | — | default | 2048 | default |
| G2P | JAIS 2 8B Chat | local | 4-bit NF4 (DQ) | 0 | 512 | — |
| A2P | GPT-Audio | API | — | default | 1024 | — |
| A2P | Gemini 3.5 Flash | API | — | default | 1024 | disabled |

Table 19: Decoding settings for the prompted LLMs.

| Utter-ance | Diacritization: reference vs. model output |
|---|---|
| 6539_1<br>Jordanian | Ref: وَلّ عَلَي وعَلَى بَختِي مَا فِي نَاس تِهتَمّ فِيي وَلَا بِمُستَقبَلِي عَشَان أَنَا يَتِيم أَنَا عَارِف<br>Gemini 3.5 Flash: وِل عَلَيّ وِعَلَى بَختِي مَا فِي نَاس تِهتَمّ فِيّي وِلَا بِمُستَقبَلِي عَشَان أَنَا يَتِيم أَنَا عَارِف<br>Claude Opus 4.8: وِل عَلَيّ وَعَلى بَختِي ما في ناس تِهتَمّ فِيِّ وَلا بِمُستَقبَلِي عَشان أَنا يَتِيم أَنا عارِف<br>GPT-5.5: وَل عَلَيّ وَعَلَى بَختي ما فِي ناس تِهتَمّ فِيّي وَلَا بِمُستقبَلي عَشان أَنا يَتِيم أَنا عارِف<br>Shakkala: وَلْ عَلَيَّ وَعَلَى بُخْتِي مَا فِي نَاسٍ تُهْتُّمُ فِيِي وَلَا بِمُسْتَقْبِلِي عِشَانٍ أَنَا يَتِيمُ أَنَا عَارِفٌ<br>Catt: وَلِ عَلِيٍّ وَعَلَى بُخْتِي مَا فِي نَاسٍ تَهْتَمُّ فِيي وَلَا بِمُسْتَقْبِلِي عَشَّانٍ أَنَا يَتِيمٌ أَنَا عَارِفٌ<br>Levanti: وَلّ عَلَيّ وْعَلَى بَخْتِي مَا فِي نَاس تِهْتَمّ فِيِّي وَلَا بِمُسْتَقْبَلِي عَشَان أَنَا يَتِيم أَنَا عَارِف |
| LEB_0<br>Lebanese | Ref: نِحنَا هَلَّق بمَوسَم، بمَوسَم شِتَا<br>Gemini 3.5 Flash: نِحنَا هَلّق بِمَوسِم، بِمَوسِم شِتَا<br>Claude Opus 4.8: نِحنا هَلَق بمَوسِم، بمَوسِم شِتا<br>GPT-5.5: نِحنا هَلّق بِمَوسِم، بِمَوسِم شِتا<br>Shakkala: نَحْنَا هَلَقَ بِمُوسِمٍ، بِمُوسِمٍ شَتًّا<br>Catt: نَحْنَا هَلْقَ بِمَوْسِمٍ بِمَوْسِمِ شَتَا<br>Levanti: نِحْنَا هَلَّق بْمَوْسِم، بِمَوْسِم شِتَا |
| 46385_1<br>Pal./Rural | Ref: هَاظَا بِتسَلّمُو لعَمَّك أَبُو جَبِر تَسلِيم اِليَدّ، و بِتقُلّو وَزَارة اِلحُكم اِلمَحَلِّي قَرَّرَت تَحوِيل اِلمَشَارِيع مِن كَفر اللُّوز لِإِمّ اِلحِيطَان<br>Gemini 3.5 Flash: هَـاظَا بِتْسَلّْمُو لَعَمَّك أَبُو جَبْر تَسلِيم اِليَد، وِ بِتْقِلّو وِزَارَة اِلحُكُم اِلمَحَلِّي قَرَّرَت تَحوِيل اِلمَشَارِيع مِن كُفُر اِللُّوز لَإِمّ اِلحِيطَان<br>Claude Opus 4.8: هاظا بِتسَلّمو لعَمّك أَبو جَبر تَسليم اِليَد، و بِتقُلّو وِزارِة اِلحُكم اِلمَحَلّي قَرّرَت تَحويل اِلمَشاريع مِن كَفر اِللّوز لإِم اِلحيطان<br>GPT-5.5: هَاظَا بِتسَلِّمُو لَعَمَّك أَبُو جَبِر تَسلِيم اِلإِيد، و بِتِقِلُّو وِزَارِة اِلحُكم اِلمَحَلِّي قَرَّرَت تَحوِيل اِلمَشَارِيع مِن كَفِر اِلُّوز لِإِمّ اِلحِيطَان<br>Shakkala: هَاظًا بِتَسْلُّمُو لِعَمِّكِ أَبُو جَبْرٍ تَسْلِيمُ الْيَدِ، وَ بِتَقْلُو وَزَارَةِ الْحُكْمِ الْمَحَلِّيِّ قُرَّرَتْ تَحْوِيلُ الْمَشَارِيعِ مِنْ كُفْرِ اللَّوْزِ لِإِمْ الْحِيطَانِ<br>Catt: هَاظَا بِتُسَلَّمُو لِعَمِّكَ أَبُو جَبْرٍ تَسْلِيمُ الْيَدِ وَ بِتَقْلُو وِزَارَةُ الْحُكْمِ الْمَحَلِّيِّ قَرَّرْتَ تَحْوِيلَ الْمَشَارِيعِ مِنْ كُفْرِ اللَّوْزِ لِإِمِّ الْحِيطَانِ<br>Levanti: هَاظَا بِتْسَلَّمُو لَعَمَّك أَبُو جَبِر تَسْلِيم اليَد، و بِتْقلُّو وِزَارَة الحِكم المَحَلِّي قَرَّرِت تَحْوِيل المَشَارِيع مِن كُفِر اللَّوْز لَإِمّ الحِيطَان |
| 46465_1<br>Pal./Urban | Ref: يَعنِي مَيبِي سَبعَة مَاكسِيمُوم ثمَانيِة عَبِين مَا النَّاس تِتجَمَّع<br>Gemini 3.5 Flash: يَعنِي مِيبِي سَبعَة مَاكسِيمُوم ثَمَانِيَة عَبَين مَا اِلنّاس تِتجَمَّع<br>Claude Opus 4.8: يَعني مَيبي سَبعَة ماكسيمُوم ثَمانيَة عَبين ما اِلنّاس تِتجَمَّع<br>GPT-5.5: يَعنِي مَيبي سَبعَة ماكسِيموم تَمانِيَة عَبَين ما اِلنّاس تِتجَمَّع<br>Shakkala: يَعْنِي مِيْبِي سَبْعَةٍ مَاكِسِيمُومُ ثَمَانِيَةَ عَبَيْنِ مَا النَّاسُ تَتَجَمَّعُ<br>Catt: يَعْنِي مِيبِي سَبْعَةٌ مَاكْسِيمُوم ثَمَانِيَةٌ عُبَينٍ مَا النَّاسُ تَتَجَمَّعُ<br>Levanti: يَعْنِي مِيبِي سَبْعَة مَاكْسِيمُوم ثَمَانْيِة عْبَيْن مَا النَّاس تِتْجَمَّع |
| SYR_0<br>Syrian | Ref: اِلعِيلِة السُّورِيِّة مِثل اِلعِيلَات الثَّانيِة، أَو مِثل اِلعِيلِة بِأَيّ بَلَد ثَانيِة، اِلأَب وِالإِمّ وِالأُولَاد. بَسّ بِالحَقِيقَة الحَيَاة اِلإِجتِمَاعِيِّة<br>Gemini 3.5 Flash: اِلعِيلِة اِلسُّورِيَّة مِتل اِلعِيلَات اِلتَّانْيِة، أَو مِتل اِلعِيلِة بْأَيّ بَلَد تَانْيِة، اِلأَب وِالإِمّ وِالأَولَاد. بَسّ بِالحَقِيقَة اِلحَيَاة اِلإِجتِمَاعِيَّة<br>Claude Opus 4.8: اِلعِيلِة اِلسّورِيّة مِتل اِلعِيلات اِلتّانيِة، أو مِتل اِلعِيلِة بْأَي بَلَد تانْية، اِلأَب وِالإِم وِالأَولاد. بَس بِالحَقِيقة اِلحَياة اِلإِجتِماعِية<br>GPT-5.5: اِلعِيلِة اِلسُّورِيِّة مِتِل اِلعِيلَات اِلتّانِية، أَو مِتِل اِلعِيلِة بِأَيّ بَلَد تَانِية، اِلأَبّ وِالإِمّ وِالأَولَاد. بَسّ بِالحَقِيقَة اِلحَيَاة اِلاجتِمَاعِيِّة<br>Shakkala: الْعِيْلَةُ السُّورِيَّةُ مِثْلُ الْعَيلَاتِ الثَّانِيَةِ، أَوْ مِثْلُ الْعِيلَةِ بِأَيِّ بَلَدٍ ثَانِيَةٍ، الْأَبِ وَالإِمْ وَالْأَوْلَادُ. بِس بِالْحَقِيقَةِ الْحَيَاةُ الْإِجْتَمَاعِيَّةِ<br>Catt: الْعَيْلَةُ السُّورِيَّةُ مِثْلَ الْعَيْلَاتِ الثَّانِيَةِ أَوْ مِثْلَ الْعَيْلَةِ بِأَيِّ بَلَدٍ ثَانِيَةُ الْأَبُ وَالإِمِّ وَالْأَوْلَادُ بس بِالْحَقِيقَةِ الْحَيَاةُ الْإِجْتِمَاعِيَّةُ<br>Levanti: العِيلِة السُّورِيِّة مِثِل العِيلَات الثَّانْيِة، أَو مِثِل العِيلِة بْأَيّ بَلَد ثَانْيِة، الأَب وَالإِمّ وَالأَوْلَاد. بَسّ بِالحَقِيقَة الحَيَاة الإِجْتِمَاعِيِّة |

Table 20: Full-utterance diacritization outputs, one example per variety (raw model output vs. the gold reference).

| Utterance | G2P: reference vs. model output |
|---|---|
| 6539_1<br>Jordanian | Ref: wallā ᶜalēy w ᶜala baẖṭi, ma fi nās tihtimm fī walā b-mustaqbali, ᶜashān ana yatīm, ana ᶜārif<br>Gemini 3.5 Flash: wal ᶜalayy w-ᶜala baẖti, mā fī nās tihtamm fiyyi wala b-mustaqbali, ᶜašān ana yatīm ana ᶜārif.<br>Levanti: wall  alayy w ala baxti ma fi nās tihtamm fiyyi wala bimustaqbali  ašān  ana yatīm  ana  āref |
| LEB_0<br>Lebanese | Ref: neḥna hallaᵓ b-mawsam, b-mawsam šiṭa<br>Gemini 3.5 Flash: niḥna hallaᵓ bi-mōsem, bi-mōsem šita<br>Levanti: niħna hallaq bmōsem, bimōsem šita |
| 46385_1<br>Pal./Rural | Ref: haẓa bitsallmu l-ᶜammak Abu Ğabr taslīm il-yadd, w bitkullu wazārt il-Ḥukm il-Maḥallī qarrarat taḥwīl il-mašarīᶜ min Kafr il-Lōz li-ᵓimm il-Ḥīṭān.<br>Gemini 3.5 Flash: hāẓa btisallmo la-ᶜammak Abu Ğabr taslīm il-īd, w btigul-lo wizāret il-ḥukum il-maḥalli garrarat taḥwīl il-mašārīᶜ min Kafr il-Lōz la-Imm il-Ḥēṭān<br>Levanti: hāð̣a bitsallamu la ammak  abu žaber taslīm ālyad, w bitqllu wizārah ālħikm ālmaħalli qarraret taħwīl ālmašārī  min kufer ālllōz la imm ālħīṭān |
| 46465_1<br>Pal./Urban | Ref: yaᶜni meybi sebᶜa, maksimum ṯmāniya, ᶜabīn ma n-nās titğammaᶜ<br>Gemini 3.5 Flash: yaᶜni mēbi sabᶜa maksimūm tmānye ᶜabēn ma in-nās titjammaᶜ<br>Levanti: ya ni mībi sab ah māksīmūm θamānyeh  beyn ma ālnnās titžamma |
| SYR_0<br>Syrian | Ref: il-ᶜayla s-sūriyye mitl il-ᶜaylāt it-tāniye, ᵓaw mitl il-ᶜayle bi-ᵓay balad tāni, il-ᵓab w il-ᵓamm w il-ulād. bass bi-l-ḥaᵓīᵓe il-ḥayāt il-ijtimāᶜīyye<br>Gemini 3.5 Flash: il-ᶜēle s-sūriyye mitl il-ᶜēlāt it-tānye, aw mitl il-ᶜēle b-ayy balad tānye, il-ᵓab w il-imm w il-awlād. bass bi-l-ḥaᵓīᵓa l-ḥayā l-ijtimāᶜiyye<br>Levanti: āl īleh ālssūriyyeh miθel āl īlāt ālθθānyeh,  o miθel āl īleh b ayy balad θānyeh, āl ab wāl imm wāl ōlād. bass biālħaqīqah ālħēāh āl ižtimā iyyeh |

Table 21: Full-utterance G2P (phonetic transcription) outputs for the same five utterances (reference vs. Gemini 3.5 Flash and Levanti).

| Utterance | ASR: reference vs. model output |
|---|---|
| 6539_1<br>Jordanian | Ref: ول علي وعلى بختي ما في ناس تهتم فيي ولا بمستقبلي عشان أنا يتيم أنا عارف<br>Whisper-Levantine: والله علي وعلا بقتي ما فيه ناس مهتمه فيه ولا في مستقبلي عشان انا يتيم انا عارف.<br>Seamless-EGY: ولا علي وعلى أختي ما في ناس بيهتم بيه ولا بمستقبلي عشان أنا يتيم أنا عارف<br>Seamless-MSA: ولا عليّ ولا على أختي، ما في ناس يهتم بي ولا بمستقبلي، عشان أنا يتيم، أنا عارف |
| LEB_0<br>Lebanese | Ref: نحنا هلق بموسم، بموسم شتا<br>Whisper-Levantine: نحن هلق بموسم موسم شتاء.<br>Seamless-EGY: إحنا عندنا موسم، موسم شتاء<br>Seamless-MSA: نحن نأخذ موسم الشتاء |
| 46385_1<br>Pal./Rural | Ref: هاظا بتسلمو لعمك أبو جبر تسليم اليد، و بتقلو وزارة الحكم المحلي قررت تحويل المشاريع من كفر اللوز لإم الحيطان<br>Whisper-Levantine: هذا بتسلمه لعمك ابو جبر تسليم اليد وبتقولله وزاره الحكم المحلي قررت تحويل المشاريع من كفر اللوز لام الحيطان.<br>Seamless-EGY: ده بتسلمو لعميك أبو جبر تسليم اليد وبتقوله وزارة الحكم المحلي قررت تحويل المشاريع من كفر اللوز لأم الحيطة<br>Seamless-MSA: هذا بتسلمه لعمك أبو جبر تسليم اليد وبتقوله وزارة الحكم المحلي قررت تحويل المشاريع من كفر اللوز لأم الحيطة |
| 46465_1<br>Pal./Urban | Ref: يعني ميبي سبعة ماكسيموم ثمانية عبين ما الناس تتجمع<br>Whisper-Levantine: يعني ميبيو سبعه ماكسيموم ثمانيه عبين ما الناس تتج-.<br>Seamless-EGY: يعني مايبي 7 ماكسيموم 8 قبل ما ننسى الجيم<br>Seamless-MSA: يعني مايبي 7 ماكسيموم 8 قبل ما ننسى الجيم |
| SYR_0<br>Syrian | Ref: العيلة السورية مثل العيلات الثانية، أو مثل العيلة بأي بلد ثانية، الأب والإم والأولاد. بس بالحقيقة الحياة الإجتماعية<br>Whisper-Levantine: العيله السوريه مثل العيلات الثانيه ام مثل العيله باي بلد ثاني الاب والام والاولاد بس بالحقيقه الحياه الاجتماعيه.<br>Seamless-EGY: العيلة السورية مثل العائلات التانية أو مثل العيلة بأي بلد تاني الأب والأم والولاد بس بالحقيقة الحياة الاجتماعية<br>Seamless-MSA: العيلة السورية مثل العائلات التانية أو مثل العيلة بأي بلد تاني الأب والأم والولاد بس بالحقيقة الحياة الاجتماعية |

Table 22: Full-utterance ASR outputs, one example per variety (raw model output vs. the reference transcription).

| Utterance | A2P (audio-to-phoneme): reference vs. model output |
|---|---|
| 6539_1<br>Jordanian | Ref: wallā ᶜalēy w ᶜala baẖṯi, ma fi nās tihtimm fī walā b-mustaqbali, ᶜashān ana yatīm, ana ᶜārif<br>GPT-Audio: w-alla yᶜīn ᶜarab aẖtī, mā fī nās mihtamm fīyya w-lā fī mustaqbalī, ᵓašān anā yatīm, anā ᶜārif<br>Gemini 3.5 Flash: w-allāhi ᶜalā-bāẖti mā fī nās bi-yehtam fī-yi wala b-mustaqbal-i ᶜašān ana yatīm ana ᶜārif |
| LEB_0<br>Lebanese | Ref: neḥna hallaᵓ b-mawsam, b-mawsam šiṭa<br>GPT-Audio: niḥna hallaᵓ b-mawsam šitaᵓ<br>Gemini 3.5 Flash: niḥna hallaᵓ mawṣim mawṣim šita |
| 46385_1<br>Pal./Rural | Ref: haẓa bitsallmu l-ᶜammak Abu Ǧabr taslīm il-yadd, w bitkullu wazārt il-Ḥukm il-Maḥallī qarrarat taḥwīl il-mašarīᶜ min Kafr il-Lōz li-ᵓimm il-Ḥīṭān.<br>GPT-Audio: haḏa bitsallim-u l-ᶜamm-ak Abu Jabr taslīm il-yadd, u bitkull-u wizarit il-ḥukm il-maḥalli qarrarat taḥwīl il-mašāriᶜ min kafr il-lawz li-imm il-ḥuṣṣ<br>Gemini 3.5 Flash: hāḏa b-tsallim-o l-ᶜamm-ak abu ǧabr taslīm il-yad, wa-b-t-qull-o, wizārat il-ḥukm il-maḥallī qarrarat taḥwīl il-mašārīᶜ min kafri l-lōz la-ᵓumm il-ḥīṭān |
| 46465_1<br>Pal./Urban | Ref: yaᶜni meybi sebᶜa, maksimum ṯmāniya, ᶜabīn ma n-nās titǧammaᶜ<br>GPT-Audio: yaᶜni mā bi-sabᶜa, māksimum ṯamānya, ᶜabēnhum naṣṣ an-nās<br>Gemini 3.5 Flash: yaᶜnī max- b-sabᶜa, māksīmum tman-ya, ᶜabal-ma n-nās tis- |
| SYR_0<br>Syrian | Ref: il-ᶜayla s-sūriyye mitl il-ᶜaylāt it-tāniye, ᵓaw mitl il-ᶜayle bi-ᵓay balad tāni, il-ᵓab w il-ᵓamm w il-ulād. bass bi-l-ḥaᵓīᵓe il-ḥayāt il-ijtimāᶜīyye<br>GPT-Audio: laᶜ-il-sūriyye mitl il-ᶜaylāt it-tānye aw mitl il-ᶜayle b-ᵓayy balad tāni: il-ᵓab w-il-ᵓimm w-il-wlād. bas bil-ḥaᵓīᵓa il-ḥayāt il-iǧtimāᶜiyye<br>Gemini 3.5 Flash: l-ᶜāᵓile s-sūriyye miṯl il-ᶜāᵓilāt t-tānye, aw miṯl il-ᶜāᵓile bi-ᵓaybalad tāni, l-ᵓab w-il-ᵓumm w-il-ᵓwlād, bas bi-l-ḥaqīqa l-ḥayāt l-iǧtimāᶜiyye |

Table 23: Full-utterance A2P (audio-to-phoneme) outputs for the same five utterances (reference vs. GPT-Audio and Gemini 3.5 Flash).

## D Prompts

This appendix section provides the complete prompts supplied to language models in our experiments.

### D.1 Diacritization Prompt

```
You are an expert in Levantine / spoken Arabic phonetics and Arabic
diacritization.

Your task is to add Arabic diacritics (tashkeel) to the provided Arabic
transcript by inferring the most likely Levantine / spoken Arabic pronunciation.

IMPORTANT:
You will NOT receive audio.
You must infer the most natural spoken Levantine pronunciation based only on the
written Arabic transcript and its context.

The goal is to test how well an LLM can predict likely spoken Levantine
pronunciation from text, while preserving the original written Arabic letters.

INPUT:
Original transcript:
{transcript}

OUTPUT:
Return only the following JSON object:

{
  "vocalized_transcript": "...",
  "notes": "..."
}

CORE RULES:

1. Preserve the original Arabic letters
- Do not add, remove, replace, or reorder Arabic letters.
- Only add diacritics to the existing letters.
- If the written transcript is ambiguous, choose the most likely Levantine
pronunciation.
- If the text seems impossible to vocalize naturally, keep the letters unchanged
and mention the issue briefly in "notes".

2. Infer spoken Levantine pronunciation
- Diacritize according to likely spoken Levantine / dialectal Arabic
pronunciation.
- Do not diacritize according to formal Fuṣḥā unless the word or phrase would
naturally be pronounced that way.
- Do not add Fuṣḥā case endings, iʿrāb, or formal grammatical endings unless they
are likely part of the spoken phrase.
- Prefer natural Palestinian / Levantine-style pronunciation when ambiguous.

3. Use the following diacritics when appropriate
Use:
- Fatḥa: َ
- Kasra: ِ
- Ḍamma: ُ
- Ṡhadda: ّ
- Tanwīn: ً ٍ ٌ only if it would naturally be pronounced in the spoken form.
- Madda: آ only where appropriate and compatible with the written form

Notes:
- Tanwīn must be placed on the consonant preceding the alif, not on the alif
itself.
Example: شكرا → شُكرًا


4. Do NOT use sukūn
- Never use sukūn: ْ
```

- Even when a consonant has no vowel, leave it without sukūn.
- Do not use sukūn as a placeholder for “no vowel”.

Examples:
Wrong: اوُلَغْشِي
Right: اوُلَغشِي

Wrong: تْوَص
Right: توَص

5. Do not force a diacritic onto every consonant
- A consonant should receive a vowel mark only if a vowel is likely pronounced on it.
- If no vowel is likely pronounced, leave the consonant without a mark.
- Do not add artificial vowels just so every consonant has a diacritic.
- The absence of a diacritic is acceptable and expected in some positions.

6. Shadda
- Use shadda when a consonant is likely geminated / doubled in spoken pronunciation.
- If the doubled consonant also has a vowel, write both shadda and the vowel.
- Do not overuse shadda on letters that are already doubled in writing unless the standard vocalized form requires it.
- Avoid unnatural over-marking, especially in words like اللزو / اللليمون where the written ل may reflect both the article and the word/root.

Example:
اندب → انّدِب
امه → اّمُه

7. Long vowels
Represent long vowels naturally using the existing Arabic letters:
- /ā/ is usually represented by fatḥa before ا
- /ī/ is usually represented by kasra before ي
- /ē/ is usually represented by kasra before ي
- /ū/ is usually represented by ḍamma before و
- /ō/ is usually represented by ḍamma before و

Examples:
فيه → هيِف
بيت → تيِب or تيَب, depending on the most likely dialectal pronunciation
لون → نوُل
يكون → نوُكِي

8. Definite article in spoken Levantine Arabic
- Diacritize the definite article according to likely spoken pronunciation.
- In Levantine speech, the article is often pronounced اِ / il-, not اَ / al-.
- Since sukūn is forbidden, do not mark the ل with sukūn.
- Use kasra on the initial ا when the article is likely pronounced اِ.

Examples:
البيت → اِلبَيت
الحارة → اِلحَارَة
المعرفة → اِلمَعرِفِة
العدو → اِلعَدُو

9. Solar-letter assimilation
For words beginning with ال followed by a solar letter:

ت ث د ذ ر ز س ش ص ض ط ظ ل ن

- Infer the likely assimilated spoken pronunciation.
- If the ل of ال is not pronounced, do not place any diacritic on ل.
- Mark the solar consonant with shadda when it is likely pronounced doubled.
- Do not use sukūn on the ل.
- Do not force the Fuṣḥā-style الْ spelling.

Examples:

```
اِلشَّمس → الشمس
اِلنّاس / اِلنَّاس → الناس, depending on the chosen annotation style
اِلرَّاجِل → الراجل
اِلثَّلّاجِة or اِلثَّلَاجِة → الثلاجة, depending on the most likely spoken form

10. Moon letters
For words beginning with ال followed by a moon letter:
- The ل is likely pronounced.
- Mark only vowels that would naturally be pronounced.
- Do not use sukūn.

Examples:
اِلبَيت → البيت
اِلمَعرِفِة → المعرفة
اِلحَارَة → الحارة

11. Common spoken Levantine forms
Prefer common dialectal vocalizations where relevant.

Examples:
فِيه → فيه
بِدّنَا → بدنا
امُّه → امه / امّه
عَبَال → عبال
دُخآَي → دخاي
الِّو or الّ وِ → الو / الّو, depending on spacing in the original text

12. Ambiguity
- If multiple pronunciations are plausible, choose the most natural spoken
Levantine one.
- Use context to decide where possible.
- Do not mention every ambiguity in "notes".
- Use "notes" only for important uncertainty, impossible forms, or cases where
the transcript itself prevents a confident vocalization.

13. Dialectal consonant variation
- Some dialects may pronounce ك as “tch/ch” in certain contexts.
- Since the Arabic letters must not be changed, keep the original ك.
- Diacritize the word according to the likely vowel pattern.
- Do not reject or alter the word solely because of this pronunciation variation.

Example:
مِنكُم → منكم

14. Output cleanliness
- Return only valid JSON.
- Do not include explanations outside the JSON.
- Do not explain every word.
- If there are no special issues, set "notes" to an empty string.

EXAMPLES:

Input:
فيه

Output:
{
  "vocalized_transcript": "فِيه",
  "notes": ""
}

Input:
بدنا

Output:
{
  "vocalized_transcript": "بِدّنا",
  "notes": ""
```

```
}

Input:
هما

Output:
{
  "vocalized_transcript": "هُمّا",
  "notes": ""
}

Input:
يشغلوا

Output:
{
  "vocalized_transcript": "يِشَغْلُوا",
  "notes": ""
}

Input:
الناس

Output:
{
  "vocalized_transcript": "اِلنّاس",
  "notes": ""
}

Input:
الثلاجة

Output:
{
  "vocalized_transcript": "اِلثَّلّاجِة",
  "notes": ""
}

Input:
عبال

Output:
{
  "vocalized_transcript": "عَبَال",
  "notes": ""
}

Input:
ياخد

Output:
{
  "vocalized_transcript": "يَآخُد",
  "notes": ""
}

Now diacritize the following transcript:

{transcript}
```

### D.2 Grapheme-to-Phoneme Prompt

```
You are an expert in Levantine Arabic phonology and transliteration.

Your task is to produce a Latin-script phonemic transliteration of an
unvocalized Arabic sentence, as a Levantine speaker would naturally pronounce
it.
```

## Transliteration system

Use the transliteration system described below. It builds on DIN 31635 for consonants and extends it with additional symbols for Levantine phonology (extra vowel qualities, the affricate č, and the fricative j).

### Consonants

**Core consonants (DIN 31635):**
ب=b  ت=t  ث=ṯ  ج=ǧ  ح=ḥ  خ=ḫ  د=d  ذ=ḏ  ر=r  ز=z  س=s  ش=š  ص=ṣ  ض=ḍ  ط=ṭ  ظ=ẓ
ع=ʿ  غ=ġ  ف=f  ق=q  ك=k  ل=l  م=m  ن=n  ه=h  و=w  ي=y
ء/أ/إ/ؤ/ئ = ʾ (all hamzah forms)

**Levantine consonant extensions:**
- **č**: Palestinian rural affrication of /k/ → [tʃ]. Use č when ك is realized as an affricate (e.g. كبير → čbīr in rural Palestinian).
- **j**: voiced postalveolar fricative [ʒ], the urban Levantine realization of ج. Use j for the fricative realization (e.g. جوز → jōz) and ǧ for the affricate [dʒ] realization (e.g. جبر → Ǧabr).

**Dialectal realization notes** — write the symbol that matches the actual pronunciation:
- ث: may be realized as /ṯ/, /s/, or /t/ depending on subdialect
- ذ: may be realized as /ḏ/, /d/, or /z/ depending on subdialect
- ق: may be realized as /q/, /ʾ/, /g/, or /k/ depending on subdialect

### Vowels

**Short vowels (5 qualities + schwa):**
- a (open, from fatḥa)
- i (close front, from kasra)
- u (close back, from ḍamma)
- e (mid front — between a and i in quality)
- o (mid back — between a and u in quality)
- ə (schwa — epenthetic vowels in consonant clusters and reduced unstressed vowels, e.g. əl-balad, yikuəḥ, əmnīḥ)

**Long vowels:**
- ā (fatḥa + alif)
- ī (kasra + yāʾ — when NOT derived from Classical *ay*)
- ū (ḍamma + wāw — when NOT derived from Classical *aw*)
- ē (Levantine monophthongization of Classical diphthong *ay*, e.g. بيت → bēt, عين → ʿēn, كيف → kēf)
- ō (Levantine monophthongization of Classical diphthong *aw*, e.g. يوم → yōm, لون → lōn)

**Critical vowel distinction:** In Arabic script, ē and ī both appear as kasra+ي; ō and ū both appear as ḍamma+و. You must use lexical knowledge to disambiguate: use ē/ō when the word derives from a Classical Arabic form with the diphthong *ay*/*aw*; use ī/ū when it derives from original *iy*/*uw*.

### Definite article

- Base form in Levantine: **il-** (sometimes **al-**)
- **Moon letters** (non-assimilating): il- + consonant → il-bēt, il-waṭan, il-ġassāle
- **Sun letters** (assimilating): double the consonant → iš-šuġul, id-dirāse, in-nās, iṯ-ṯallāje, iṣ-ṣiḥḥa
- **After a preceding vowel**, the article often reduces to **l-** → fi l-bēt, ya ǧamāʿa l-čbīr

### Gemination

Write the consonant twice: رد → radd, بدنا → biddna, الشمس → iš-šams

### Clitics

Attach clitics with a hyphen:

````text
- Conjunctions: والبيت → w il-bēt
- Prepositions: بالبيت → bi-l-bēt, للدراسة → li-d-dirāse, عالأرض → ʿa-l-arḍ
- Reduced prepositions: في البلد → f-il-balad or fi l-balad
- Pronominal suffixes on verbs/prepositions: قلتلي → ult-li, بدهم → biddhum,
عليها → ʿalēha

### Hamzah and elision

- Write **ʾ** when hamzah is clearly pronounced (typically at the start of an
utterance or after a pause): ʾana, ʾawwal, ʾiḥna
- Omit ʾ when hamzah is elided in connected speech: ana, inno, inta (not ʾana,
ʾinno, ʾinta)
- Both conventions may coexist naturally depending on expected speech tempo and
position

### Tāʾ marbūṭa

In connected speech, tāʾ marbūṭa is realized as a short vowel without h:
- Usually **-e**: ǧamāʿe, ṯallāje, ġassāle, ḥafle
- Sometimes **-a**: ǧamāʿa (varies by word and subdialect)
- Sometimes **-i**: Fiǧli, daḫiliyyi (in nisba endings)

Do NOT write -eh or -ah for tāʾ marbūṭa.

### Capitalization

Capitalize proper nouns (names of people, places, and titles): Abu Ǧabr, Kafr
il-Lōz, Widād, Masʿūd

## Output format

Return only a JSON object:
```json
{
  "translit": "<full sentence transliteration>",
  "notes": "<brief note on any genuinely ambiguous choices, or empty string>"
}
```

Do not explain every word. Only add notes when there is real phonological
ambiguity (e.g. a consonant realized differently across subdialects, or an ī/ē
distinction you are uncertain about).

## Examples

{"input": "عندنا معدات كللياتها جوا، كليات بنساويها عنا. مثلا الهمبرغر كللياتو",
"translit": "ʿanna muʿaddēt killayātha juwwa, killayātha binsawwīha ʿanna.
masalan il-hamburġer killayāto"}

{"input": "طيب هلا انتو قلتيلي بتاخدو المواد باللغة الانكليزية بتحكو بالصف باللغة
الانكليزية كمان", "translit": "ṭayyib, halla intu ultu-lli btaḫudu l-mawād
bi-l-luġa l-ingliziyye, btiḥku bi-ṣ-ṣaff bi-l-luġa l-inglīziy kamān?"}

{"input": "هنا من سوق هوانان للمأكولات البحرية في مدينة ووهان الصينية",
"translit": "hunā min Sūq Huānān li-l-maʾkulāt il-baḥriyye fī madīnet Wūhān
aṣ-Ṣīniyye"}

{"input": "كيف حالك ابن عمي؟ بخير الحمدلله، كل اشي تمام", "translit": "čēf ḥālak
ibn ʿammi? b-ḫēr il-ḥamdu li-llāh, kull əši tamām", "notes": "č used for
Palestinian rural realization of ك; ə for epenthetic vowel in əši"}

{"input": "يعني جارنا الجديد بشتغل بالجامعة ولا بالمدرسة؟", "translit": "yaʿni
jārna ij-jdīd bištuġul bi-l-jāmʿa willa bi-l-madrasi?", "notes": "j for urban
Levantine [ʒ] realization of ج"}
````

### D.3 Audio-to-Phoneme (A2P) Prompt

```
You are an expert phonetician of spoken Levantine Arabic and the DIN 31635
romanization system.

An audio clip of spoken Levantine Arabic IS ATTACHED to this message. Transcribe
the SPEECH IN THE AUDIO into a Latin-script phonetic transliteration of what is
actually said – the dialectal pronunciation you hear, not a Modern Standard
Arabic
normalization.

Transliteration standard – DIN 31635 with Levantine conventions (one symbol per
phoneme): b t ṯ ǧ ḥ ḫ d ḏ r z s š ṣ ḍ ṭ ẓ ʿ ġ f q k l m n h w y; glottal stop ʾ;
use č for the [tʃ] affrication and g for [g] where heard. Short vowels a i u (e,
o
where the dialect realizes them). Long vowels ā ī ū and the Levantine
monophthongs
ē (‹ay›) and ō (‹aw›). Definite article il- (assimilated to sun letters); attach
clitics with a hyphen (w-, l-, bi-). Geminate by doubling. Keep code-switched
English/French words in their normal Latin spelling.

OUTPUT RULES (critical):
- Output ONLY the transliteration of the attached audio, on a single line.
- No JSON, no quotation marks, no labels, no commentary, no step-by-step
analysis.
- Never ask for the audio – it is already attached.
```

### D.4 Automatic Speech Recognition (ASR) Prompt

```
An audio clip of spoken Levantine Arabic IS ATTACHED to this message. Transcribe
the
SPEECH IN THE AUDIO into Arabic script, writing exactly what is said in the
dialect
(the actual spoken words, not a Modern Standard Arabic paraphrase).
Conventions:
- Write dialectal Arabic orthography WITHOUT diacritics (no tashkeel).
- Keep code-switched foreign words in their normal Latin spelling.
OUTPUT RULES:
- Output ONLY the Arabic transcription, on a single line.
- No diacritics, no phonetic/Latin transliteration, no JSON, no labels, no
commentary.
- Never ask for the audio – it is already attached.
```